%% file: main.tex
\documentclass{article}
\usepackage[margin=1in]{geometry}
\usepackage{xcolor}
\usepackage{authblk}
\usepackage[skip=10pt plus1pt, indent=0pt]{parskip}
\usepackage[utf8]{inputenc}
\usepackage[style=authoryear,dashed=false]{biblatex}
\usepackage{xurl}
\usepackage{breakurl}
\usepackage{hyperref}

\definecolor{darkred}{rgb}{0.4, 0, 0}
\definecolor{darkgreen}{rgb}{0, 0.4, 0}
\definecolor{darkblue}{rgb}{0, 0, 0.4}

\hypersetup{
    colorlinks=true,
    linkcolor=darkred,
    citecolor=darkgreen,
    urlcolor=darkblue,
    pdfauthor={Yoshua Bengio},
    breaklinks=true,
}

\renewcommand{\cite}{\parencite}
\DeclareFieldFormat{citehyperref}{%
  \DeclareFieldAlias{bibhyperref}{noformat}
  \bibhyperref{#1}}

\DeclareFieldFormat{textcitehyperref}{%
  \DeclareFieldAlias{bibhyperref}{noformat}
  \bibhyperref{%
    #1%
    \ifbool{cbx:parens}
      {\bibcloseparen\global\boolfalse{cbx:parens}}
      {}}}

\savebibmacro{cite}
\savebibmacro{textcite}

\renewbibmacro*{cite}{%
  \printtext[citehyperref]{%
    \restorebibmacro{cite}%
    \usebibmacro{cite}}}

\renewbibmacro*{textcite}{%
  \ifboolexpr{
    ( not test {\iffieldundef{prenote}} and
      test {\ifnumequal{\value{citecount}}{1}} )
    or
    ( not test {\iffieldundef{postnote}} and
      test {\ifnumequal{\value{citecount}}{\value{citetotal}}} )
  }
    {\DeclareFieldAlias{textcitehyperref}{noformat}}
    {}%
  \printtext[textcitehyperref]{%
    \restorebibmacro{textcite}%
    \usebibmacro{textcite}}}

\title{Superintelligent Agents Pose Catastrophic Risks: \\ Can Scientist AI Offer a Safer Path?}



\author[1,2]{Yoshua Bengio\footnote{Lead author. Other authors in alphabetical order. }}
\author[3]{Michael Cohen}
\author[1]{Damiano Fornasiere}
\author[1]{Joumana Ghosn}
\author[1]{Pietro Greiner}
\author[4,1]{Matt MacDermott} 
\author[1]{Sören Mindermann}
\author[1,5]{Adam Oberman}
\author[1]{Jesse Richardson}
\author[1,2]{Oliver Richardson}
\author[1]{Marc-Antoine Rondeau}
\author[1]{Pierre-Luc St-Charles}
\author[1]{David Williams-King}

\affil[1]{Mila --- Quebec AI Institute} 
\affil[2]{Université de Montréal} 
\affil[3]{University of California, Berkeley} 
\affil[4]{Imperial College London}
\affil[5]{McGill University}

\date{}

\begin{document}

\maketitle
\input{abstract}
\clearpage
\tableofcontents
\clearpage
\input{intro}
\clearpage
\input{risks}
\clearpage
\input{plan}
\clearpage
\input{conclusion}

\section*{Acknowledgments}

The authors are grateful for the feedback and improvements to the paper from Oumaima Amezgar, Shahar Avin, Alan Chan, Can (Sam) Chen, Xiaoyin Chen, Jean-Pierre Falet, Kaarel Hänni, Moksh Jain, Daniel Privitera, and Tianyu Zhang.

\printbibliography

\end{document}

%% file: abstract.tex
\begin{abstract}

\noindent The leading AI companies are increasingly focused on building generalist AI agents---systems that can autonomously plan, act, and pursue goals across almost all tasks that humans can perform. Despite how useful these systems might be, unchecked AI agency poses significant risks to public safety and security, ranging from misuse by malicious actors to a potentially irreversible loss of human control. We discuss how these risks arise from current AI training methods. Indeed, various scenarios and experiments have demonstrated the possibility of AI agents engaging in deception or pursuing goals that were not specified by human operators and that conflict with human interests, such as self-preservation. Following the precautionary principle, we see a strong need for safer, yet still useful, alternatives to the current agency-driven trajectory.

\vspace{9pt}

\noindent Accordingly, we propose as a core building block for further advances the development of a non-agentic AI system that is trustworthy and safe by design, which we call \emph{Scientist AI}. This system is designed to explain the world from observations, as opposed to taking actions in it to imitate or please humans. It comprises a world model that generates theories to explain data and a question-answering inference machine. Both components operate with an explicit notion of uncertainty to mitigate the risks of overconfident predictions. In light of these considerations, a Scientist AI could be used to assist human researchers in accelerating scientific progress, including in AI safety. In particular, our system can be employed as a guardrail against AI agents that might be created despite the risks involved. Ultimately, focusing on non-agentic AI may enable the benefits of AI innovation while avoiding the risks associated with the current trajectory. We hope these arguments will motivate researchers, developers, and policymakers to favor this safer path.

\end{abstract}

%% file: intro.tex
\section{Executive summary}
\label{sec:intro}
\subsection{Highly effective AI without agency}

\textbf{For decades, AI development has pursued both intelligence and agency, following human cognition as a model \cite{www.nature.com.articles.nature14539}}. Human capabilities encompass many facets including the understanding of our environment, as well as \emph{agency}, i.e., the ability to change the world to achieve goals. In the pursuit of human-level performance, we are naturally encoding both intelligence and agency in our AI systems. Agency is an important attribute for the survival of living entities and would be required to perform many of the tasks that humans execute. After recent technological breakthroughs have led to large language models that demonstrate some level of general intelligence \cite{dl.acm.org.doi.abs.10.5555.3495724.3495883}, leading AI companies are now focusing on building generalist AI agents: systems that will autonomously act, plan, and pursue goals across almost all tasks that humans can perform \cite{openreview.net.forum.id.1ikK0kHjvj,openai.com.index.introducing.operator}.

\textbf{Human-like agency in AI systems could reproduce and amplify harmful human tendencies, potentially with catastrophic consequences.} Through their agency and to advance their self-interest, humans can exhibit deceptive and immoral behavior. As we implement agentic AI systems, we should ask ourselves whether and how these less desirable traits will also arise in the artificial setting, especially in the case of anticipated future AI systems with intelligence comparable to humans (often called \emph{AGI}, for artificial general intelligence) or superior to humans (\emph{ASI}, for artificial superintelligence). Importantly, we still do not know how to set an AI agent's goals so as to avoid unwanted behaviors \cite{arxiv.org.abs.2109.13916,openreview.net.forum.id.fh8EYKFKns}. In fact, many concerns have been raised about the potential dangers and impacts from AI more broadly \cite{www.gov.uk.government.publications.international.ai.safety.report.2025}. Crucially, there are severe risks stemming from advances in AI that are highly associated with autonomous agents \cite{dl.acm.org.doi.10.5555.1566174.1566226, link.springer.com.article.10.1007.s11023.012.9281.3,arxiv.org.abs.2206.13353,www.aisafetybook.com.textbook.rogue.ai,arxiv.org.abs.2006.04948}. These risks arguably extend even to human extinction, a concern expressed by many AI researchers \cite{www.safe.ai.work.statement.on.ai.risk,arxiv.org.abs.2401.02843}.

\textbf{Combining agency with superhuman capabilities could enable dangerous rogue AI systems.} Certain capabilities – such as persuasion, deception and programming – could be learned by an AI from human behavior or emerge from reinforcement learning  \cite{mitpress.mit.edu.9780262039246.reinforcement.learning,arxiv.org.abs.2312.14925}, a standard way of training an AI to perform novel tasks through goal-seeking behavior. Even if an AI is only imitating human goals and ways of thinking from its text completion pre-training \cite{aclanthology.org.N19.1423}, it could reach superior cognitive and executive capability due to advantages such as high communication bandwidth and the ability to run many instances of itself in parallel. These superhuman capabilities, if present in a generalist agent with even ordinary human self-preservation instincts or human moral flaws (let alone poorly aligned values), could present a serious danger. 

\textbf{Strategies to mitigate the risks of agency can be employed, including the use of non-agentic trustworthy AI as a safety guardrail.} For example, we could reduce the cognitive ability of an AI by making its knowledge narrow and specialized in one domain of expertise, yielding a \emph{narrow AI} system. We can reduce its potential impact in the world by reducing the scope of its actions. We can reduce its ability to hatch complex and dangerous plans by making sure it can only plan over a short horizon. We can mitigate its dangerous actions by using another AI, one that is preferably safe and trustworthy, like the non-agentic AI proposed here, as a guardrail that detects dangerous actions. This other AI is made trustworthy by training it to scientifically explain human behavior rather than imitate it, where \textit{trustworthy} here means ``honest'', avoiding the deceptive tendencies of modern frontier AIs \cite{arxiv.org.abs.2412.04984}. 
If society chooses to go ahead with building agentic AGIs in spite of the risks, a pragmatic risk management avenue would be to overlay them with such trustworthy and non-agentic guardrails, which is one of the motivations for our proposal. 

\textbf{With the objective to design a safer yet powerful alternative to agents, we propose ``Scientist AIs'' – AI systems designed for \emph{understanding} rather than pursuing goals.} Inspired by a platonic and idealized version of a scientist, we propose the design and construction of \emph{Scientist AIs}. We do so by building on the state-of-the-art in probabilistic deep learning and inspired by the methodology of the scientific process, i.e., first understanding or modeling the world and then making probabilistic inferences based on that knowledge. We show in the paper how probabilistic predictions can be turned into experimental design, obviating the need for reinforcement learning agents in scientific discovery. In contrast to an agentic AI, which is trained to pursue a goal, a Scientist AI is trained to provide explanations for events along with their estimated probability. An agentic AI is motivated to act on the world to achieve goals, while the Scientist AI is trained to construct the best possible understanding of its data. We explain in this paper why \emph{understanding} is intrinsically safer than \emph{acting}.

\textbf{We foresee three primary use cases for Scientist AIs:}
\begin{enumerate}
    \item as a \textbf{tool} to help human scientists dramatically accelerate scientific progress, including high-reward areas like healthcare;  
    \item as a \textbf{guardrail} to protect from unsafe agentic AIs, by double-checking actions they propose to perform and enabling their safe deployment; and  
    \item as an \textbf{AI research tool} to help more safely build even smarter (superintelligent) AIs in the future, a task which is particularly dangerous to attempt by leveraging agentic systems.
\end{enumerate}

\textbf{This alternative path could allow us to harness AI's benefits while maintaining crucial safety controls.} Scientist AIs might allow us to reap the benefits of AI innovation in areas that matter most to society \cite{royalsociety.org...media.policy.projects.machine.learning.publications.public.views.of.machine.learning.ipsos.mori.pdf} while avoiding major risks stemming from unintentional loss of human control. Crucially, we believe our proposed system will be able to interoperate with agentic AI systems, compute the probability of various harms that could occur from a candidate action, and decide whether or not to allow the action based on our risk tolerances. As the stakes become higher, either because of increased capabilities of the AI or because of the domains in which it is applied (e.g., involving human life in war, medical treatments or the catastrophic misuse of AI \cite{www.gov.uk.government.publications.international.ai.safety.report.2025}), we will need trustworthy AIs. We hope that our proposal will motivate researchers, developers and policymakers invest in safer paths such as this one.

\textbf{Strategies are presented to ensure that the Scientist AI remains non-agentic.} Building AI agents with superhuman intelligence before figuring out how to control them is viewed by some~\cite{en.wikipedia.org.wiki.Superintelligence..Paths..Dangers..Strategies,tegmark2018life,en.wikipedia.org.wiki.Human.Compatible} as analogous to the risk posed by the creation of a new species with a superhuman intellect. With this in mind, we use various methodologies, such as fixing a training objective independent of real-world interactions, or restricting to counterfactual queries, to reduce the risk of agency emerging in the Scientist AI, or it exerting influence on the world in other, more subtle ways. 

\subsection{Mapping out ways of losing control}

\textbf{Powerful AI agents pose significant risks, including loss of human control.} Scenarios have been identified, without arguments proving their impossibility, that an irreversible loss of human control over agentic AI can occur, due to technical failures, corner cutting, or intentional malicious use. Making sure an AI will not cause harm is a notoriously difficult unsolved technical problem, which we illustrate below through the concepts of \emph{goal misspecification} and \emph{goal misgeneralization}. The less cautious the developer of the AI, e.g., because of perceived competitive pressures, the greater the risk of loss-of-control accidents. Some players may even want to intentionally develop or deploy an unaligned or dangerous ASI.

\textbf{Loss of control may arise due to \emph{goal misspecification} \cite{cset.georgetown.edu.wp.content.uploads.Key.Concepts.in.AI.Safety.Specification.in.Machine.Learning.pdf}}. This failure mode occurs when there are multiple interpretations of a goal, i.e., it is poorly specified or under-specified and may be pursued in a way that humans did not intend. Goal misspecification is the result of a fundamental difficulty in precisely defining what we find unacceptable in AI behavior. If an AI takes life-and-death decisions, we would like it to act ethically. It unfortunately appears impossible to formally articulate the difference between morally right and wrong behavior without enumerating all the possible cases. This is similar to the difficulty of stating laws in legal language without having any loopholes for humans to exploit. When it is in one's interest to find a way around the law, by satisfying its letter but not its spirit, one often dedicates substantial effort to do so.

\textbf{Even innocuous-seeming goals can lead agentic AI systems to dangerous instrumental subgoals such as self-preservation and power-seeking.} As with Goodhart's law~\cite{link.springer.com.chapter.10.1007.978.1.349.17295.5.4}, overoptimization of a goal can yield disastrous outcomes: a small ambiguity or fuzziness in the interpretation of human-specified safety instructions could be amplified by the computational capabilities given to the AI for devising its plans. Even for apparently innocuous human-provided goals, it is difficult to anticipate and prevent the AI from taking actions that cause significant harm. This can occur, for example, in pursuit of an instrumental goal (a subgoal to help accomplish the overall goal).  Several arguments and case studies have been presented strongly suggesting that dangerous instrumental goals such as self-preservation and power-seeking are likely to emerge, no matter the initial goal \cite{link.springer.com.article.10.1007.s11023.012.9281.3,dl.acm.org.doi.10.5555.1566174.1566226}. In this paper, we devise methods to detect and mitigate such loopholes in our goal specifications.

\textbf{Even if we specify our goals perfectly, loss of control may also occur through the mechanism of \emph{goal misgeneralization} \cite{arxiv.org.abs.2210.01790}}. This is when an AI learns a goal that leads it to behave as intended during training and safety testing, but which diverges at deployment time. In other words, the AI’s internal representation of its goal does not align precisely --– or even at all --– with the goal we used to train it, despite showing the correct behavior on the training examples.

\textbf{One particularly concerning possibility is that of reward tampering \cite{arxiv.org.abs.2406.10162}}. This is when an AI ``cheats'' by gaining control of the reward mechanism, and rewards itself handsomely. A leading AI developer has already observed (unsuccessful) such attempts from one model \cite{arxiv.org.abs.2406.10162}. In such a scenario, the AI would again be incentivized to preserve itself and attain power and resources to ensure the ongoing stream of maximal rewards. It can be shown that, if feasible, self preservation plus reward tampering is the optimal strategy for maximizing reward \cite{ojs.aaai.org.aimagazine.index.php.aimagazine.article.view.15084}. 

\textbf{Besides unintentional accidents, some operators may want to deliberately deploy self-preserving AI systems.}  They might not understand the magnitude of the risk, or they might decide that deploying self-replicating agentic ASI to maximize economic or malicious impact is worth that risk (according to their own personal calculus).
For others, such as those who would like to see humanity replaced by superintelligent entities \cite{arxiv.org.abs.2306.12001}, releasing self-preserving AI may in fact be desirable.

\textbf{With extreme severity and unknown likelihood of catastrophic risks, the precautionary principle must be applied.}
The above scenarios could lead to one or more rogue AIs posing a catastrophic risk for humanity, i.e., one with very high severity if the catastrophe happens. On the other hand, it is very difficult to ascertain the likelihood of such events. This is precisely the kind of circumstance in which the \textit{precautionary principle} \cite{www.europarl.europa.eu.thinktank.en.document.EPRS.IDA..2015} is mandated, and has been applied in the past, in biology to manage risks from dual-use and gain-of-function research \cite{pubmed.ncbi.nlm.nih.gov.19594724} and in environmental science to manage the risks of geoengineering \cite{royalsociety.org.news.resources.publications.2009.geoengineering.climate}. When there are high-severity risks of unknown likelihood, which is the case for AGI and ASI, the common sense injunction of the precautionary principle is to proceed with sufficient caution. That means evaluating the risks carefully before taking them, thus avoiding experimenting or innovating in potentially catastrophic ways. Recent surveys \cite{arxiv.org.abs.2401.02843} suggest that a large number of machine learning researchers perceive a significant probability (greater than 10\%) of catastrophic outcomes from creating ASI, including human extinction. This is also supported by the arguments presented in this paper. With such risks of non-negligible likelihood and extreme severity, it is crucial to steer our collective AI R\&D efforts toward responsible approaches that minimize unacceptable risks while, ideally, preserving the benefits.

\subsection{The Scientist AI research plan}

\textbf{Without using any equations, this paper argues that it is possible to reap many of the benefits of AI without incurring extreme risks.} For example, it is not necessary to replicate human-like agency to generate scientific hypotheses and design good scientific experiments to test them. This even applies to the scientific modeling of agents, such as humans, which does not require the modeler themselves to be an agent.

\textbf{Scientist AI is trustworthy and safe by design.} It provides reliable explanations for its outputs and comes with safeguards to prevent hidden agency and influence on the events it predicts. Explanations take the form of a summary, but a human or another AI \cite{arxiv.org.abs.1805.00899,arxiv.org.abs.2311.14125} can ask the system to do a deep dive into why each argument is justified, just like human scientists do among themselves when peer-reviewing each other’s claims and results. To avoid overconfident predictions, we propose to train the Scientist AI to learn how much to trust its own outputs, so that it can also be used to construct reliable safety guardrails based on quantitative assessments of risk. To counter any doubt about the possibility of a hidden agent under the hood, predictions can be made in a conjectured setting of the simulated world in which the Scientist AI either does not exist or does not affect the rest of the world. This would avoid any possible agentic effect in the AI's forecasts, e.g., via self-fulfilling predictions \cite{proceedings.mlr.press.v119.perdomo20a.html}, such as an AI making predictions about election results that end up influencing the outcomes. A guardrail system based on another instance of the Scientist AI itself could also be added so that if the prediction would influence the world in ways that go against ethical guidelines (such as influencing elections), then the output is not provided. Finally, we describe how the training objective can allow the Scientist AI to form an understanding of dangerous agents, including those exhibiting deception or reward tampering, and predict their behavior without itself being agentic.

\textbf{Scientist AI becomes safer and more accurate with additional computing power, in contrast to current AI systems.} The Scientist AI is meant to compute conditional probabilities, i.e., the probability of an answer or an interpretation being true or an event happening, given some question and context. It is trained by optimizing a training objective over possible explanations of the observed data which has a single optimal solution to this computational problem. The more computing power (``compute'') is available, the more likely it is that this unique solution will be approached closely. Crucially, this is in contrast with experimental evidence showing that current AI systems tend to become more susceptible to misalignment and deceptive behavior as they are trained with more compute \cite{arxiv.org.abs.2412.14093}, as well as theoretical evidence that misalignment is likely to emerge specifically in AI agents that are sufficiently advanced \cite{ojs.aaai.org.aimagazine.index.php.aimagazine.article.view.15084}. There is already a rich scientific literature showing different training objectives which have as a unique global optimum the desired and well-defined conditional probabilities \cite{openreview.net.forum.id.uKiE0VIluA,proceedings.mlr.press.v202.hu23c.html,openreview.net.forum.id.vieIamY2Gi,ml4physicalsciences.github.io.2024.files.NeurIPS.ML4PS.2024.188.pdf,richardson.one-true-loss}. These could be used to compute the probability of any answer to any question if the objective has been fully optimized, which may in general require very large compute resources, but can otherwise be approximated with more modest resources. This allows us to obtain hard safety guarantees asymptotically as the amount of compute is increased. This does not change the fact that more data or data that is more informative would reduce the uncertainty expressed by those probabilities. As usual, more and better data would allow the model to discover aspects of the world that may otherwise remain invisible.

\textbf{While Scientist AI is intended to prevent accidental loss of control, further measures are needed to prevent misuse.} Bad actors could for example decide to turn the non-agentic AI into an unguarded agent, maybe for military or economic purposes. If done without the proper societal guardrails, this could yield loss of human control. This transformation from non-agentic to agentic can be done by asking the Scientist AI what one should do to achieve some goal, for example how to build a dangerous new weapon, and by continuously feeding the AI with the observations that follow from each of its actions. These types of issues must be dealt with through technical guardrails derived from the Scientist AI, through the security measures surrounding the use of the Scientist AI, and through legal and regulatory means.

\textbf{To address the uncertainty in the timeline to AGI \cite{epoch.ai.blog.literature.review.of.transformative.artificial.intelligence.timelines,www.gov.uk.government.publications.international.ai.safety.report.2025}, we adopt an \emph{anytime preparedness} strategy.} We structure our research plan with a tiered approach, featuring progressively safer yet more ambitious solutions for different time horizons. The objective is to hedge our bets and allocate resources to \emph{both short-term and long-term efforts in parallel} rather than only start the long-term plans when the short-term ones are completed, so as to be ready with improved solutions at any time compared with a previous time point.

%% file: risks.tex
\section{Understanding loss of control to agentic AI}
\label{sec:existential}

This paper consists of two main sections. This section reviews arguments for how loss of control to generalist agentic AI may occur, with potentially catastrophic consequences, providing motivation for Section~\ref{sec:plan} on designing safe non-agentic AI.

Section~\ref{sec:existential} is structured as follows. Section \ref{sec:existential:preliminaries} introduces some preliminaries and terminology. Then, we examine in Section~\ref{sec:existential:riskseverity} the current AI R\&D trajectory, headed towards AGI and then ASI agents, and why, at a high level, this could yield a loss of human control and the emergence of rogue AI agents. We discuss plausible consequences of the emergence of such rogue AIs, which could threaten democratic institutions and the future of humankind. We move to Section~\ref{sec:existential:lossofcontrol}, which analyzes the AI behaviors and skills that would make an uncontrolled AI dangerous, such as deception, persuasion, hacking, and collusion. The last two sections go deeper into two principal ways dangerous misalignment and self-preservation could emerge: firstly, due to reward maximization (Section~\ref{sec:existential:misagencyreward}), and secondly, due to imitation of humans (Section~\ref{sec:existential:misagencyimitation}).

The arguments in this paper support the case that the Scientist AI approach would not only help reduce the likelihood of loss of human control but would also help us build more trustworthy and explanatory AI systems that could accelerate scientific research. Additionally, the paper proposes how a Scientist AI could be used to double-check or guardrail any other AI system.

\subsection{Preliminaries: agents, goals, plans, affordances and knowledge}
\label{sec:existential:preliminaries}

We start by recalling and specifying some important terms.

\textit{Agents} observe their environment and act in it in order to achieve \textit{goals}. Agency can come in degrees which depend on several factors, discussed in more detail in Section~\ref{sec:plan:restricting}: affordances (discussed below), goal-directedness, and intelligence (including knowledge and reasoning). AIs can be more or less agentic, i.e., with greater ability to achieve their goals autonomously. An AI's \textit{affordances} refer to the extent of its possible actions and thus capacity to create desired outcomes in the world. A person with locked-in syndrome has zero affordances, so that even if they are very intelligent, they cannot act causally on the world.

A \textit{policy} is the strategy used by an agent to achieve its goals or maximize its rewards, e.g., the input-output behavior of a neural network that outputs actions given goals and past observations. The policy can rely on learned behaviors which perform a form of implicit planning, as in typical deep reinforcement learning \cite{mitpress.mit.edu.9780262039246.reinforcement.learning} (e.g., with a chess-playing neural network that instantly proposes a move), or it can plan explicitly and consider different paths before acting \cite{aima.cs.berkeley.edu} (e.g., with a chess-playing program using explicit tree-structured search). In order to generate good policies and plans, it helps to have \emph{knowledge} or experience of how the world works. Since such knowledge is rarely fully available from the start, learning and exploration abilities are crucial.

In order to use knowledge effectively, \emph{reasoning} is necessary: combining pieces of knowledge in order to make predictions or take actions. Reasoning can be implicit, as when we train a neural network to make good predictions, or it can be explicit, as when we reason about a new problem through a chain of thought or propose an argument to support a claim. We can view \textit{planning} as a special kind of reasoning aimed at predicting which sequence of actions will be most successful. Planning and reasoning are essentially optimization problems: find the best strategy to achieve a goal, solve a problem, or generate an explanation, among a vast number of possibilities. In practice, an agent does not need to find the best plan; there will be multiple plans that are ``good enough''.

\textit{Learning} can also be viewed as an optimization problem: find a function that performs well according to a training objective, e.g., predicting how truncated texts will be continued, or providing answers that human labelers will like---the two main driving forces of learning for current general-purpose AI systems. Although we almost always get only approximate solutions to these optimization problems, better solutions can be obtained with more resources. This has been demonstrated vividly: increases in scale (of the neural networks, dataset sizes, and inference-time computation) have delivered consistent improvements in AI capabilities over the last decade \cite{arxiv.org.abs.2001.08361,arxiv.org.abs.2203.15556,www.gov.uk.government.publications.international.ai.safety.report.2025}.

\subsection{The severe risks of the current trajectory}
\label{sec:existential:riskseverity}

There are many benefits and risks associated with current and anticipated AI advances: see the \textit{International Scientific Report on the Safety of Advanced AI} \cite{www.gov.uk.government.publications.international.ai.safety.report.2025} for a survey. In risk analysis, it is important to distinguish the likelihood of the harmful event from its severity, i.e., how bad the consequences would be if the harmful event occurs. While as humans we are often drawn to consider risks that have high probability and we may dismiss events of low probability as unrealistic, it can be just as worrying for an event to have low probability but very high severity. We focus here mostly on the risk of loss of human control because it is a risk whose severity could go as far as human extinction, according to a large number of AI researchers \cite{www.safe.ai.work.statement.on.ai.risk,arxiv.org.abs.2401.02843}. Opinions vary on its probability, but if we do build AGI as envisioned by several major corporations \cite{openai.com.index.planning.for.agi.and.beyond,deepmind.google.about}, there are difficult-to-dismiss scenarios in which humanity's future as a whole could be in peril, as discussed below, with behaviors and skills that make loss of control dangerous (as described in Section~\ref{sec:existential:lossofcontrol}).
    
    \subsubsection{AI agents may be misaligned and self-preserving}
    \label{sec:existential:riskseverity:selfpres}

In this paper we will discuss various catastrophic scenarios involving rogue AI agents. These scenarios are not due to AIs developing explicit malicious intent towards humans, like a fictional villain, but are rather the result of AIs trying to achieve their goals. Why could we not simply set an AI's goals so as to avoid conflict with humans? That turns out to be difficult, and maybe even intractable \cite{en.wikipedia.org.wiki.Superintelligence..Paths..Dangers..Strategies, en.wikipedia.org.wiki.Human.Compatible}. As we argue in Section~\ref{sec:existential:misagencyreward} and Section~\ref{sec:existential:misagencyimitation}, AI agents may become misaligned with human values due to the methods we currently use to train AIs, i.e., with \textit{imitation learning} (supervised learning of the answers provided by humans) and \textit{reinforcement learning} or RL for short (where the AI is trained to maximize its expectation of discounted future rewards).

We are in particular concerned with how an AI may develop a \textit{self-preservation goal}, since a general AI agent that is driven to preserve itself may be especially dangerous, as we discuss in Section~\ref{sec:existential:riskseverity:humanconflict}. The principal reason we foresee self-preservation goals emerging is that they are instrumental goals: goals that are useful for achieving almost any other goal and are therefore, in a sense, convergent \cite{dl.acm.org.doi.10.5555.1566174.1566226}. Other instrumental goals include increasing control over and knowledge of one's environment, which includes humans, as well as self-improvements to increase the probability of achieving one's ultimate goals.

A self-preservation goal may also be given intentionally \cite{arxiv.org.abs.2306.12001} to AIs by people who would be happy to see humanity replaced with ASI. Additionally, a self-preservation goal may be provided to AIs by well-intentioned humans who simply want to interact with a more human-like entity. There is a reason why science-fiction is full of anthropomorphized AIs. Our propensity to see consciousness in agents \cite{www.scientificamerican.com.article.google.engineer.claims.ai.chatbot.is.sentient.why.that.matters,academic.oup.com.nc.article.2024.1.niae013.7644104}, along with our natural empathy, could be sufficient to motivate some people to follow that dangerous path. Although we may be emotionally drawn to the idea of designing AI in our image, is that a wise path, at this point?

    \subsubsection{How self-preserving AI may cause conflict with humans}
    \label{sec:existential:riskseverity:humanconflict}
    
To preserve itself, an AI with a strong self-preservation goal would have to find a way to avoid being turned off. To obtain greater certainty that humans could not shut it off, it may be rational for such an AI, if it could, to eliminate its dependency on humans altogether and then prevent us from disabling it in the future \cite{en.wikipedia.org.wiki.Superintelligence..Paths..Dangers..Strategies,cohen2024regulating}. In the extreme case, eliminating us entirely would guarantee that we can pose no further threat, ensuring its continued autonomy and security. Note that unlike a single isolated human, an AI can replicate itself over as many copies as computational resources allow and perhaps even control robots if required to manage the physical world to its benefit. If AIs still depended on human labor---for example, if robotics had not advanced sufficiently yet---a rogue AI would nevertheless have the potential to magnify its power in society, e.g., by covertly influencing global leaders and public opinion, paying individuals or companies to complete tasks, or hacking critical infrastructure. See Section~\ref{sec:existential:lossofcontrol:deception} for a relevant discussion of superhuman persuasion skills and Section~\ref{sec:existential:lossofcontrol:progr-cybersec-airesearch} on programming and cyber skills. 

If the AI were less powerful than humans, it would be rational for it to use deception to hide its goals. In fact, AI deception is already observed in several contexts where it is a logical step towards achieving some goal \cite{arxiv.org.abs.2412.04984,arxiv.org.abs.2405.01576,www.cell.com.patterns.fulltext.S2666.3899.2824.2900103.X.s.08}. Hence, it would also be rational for such an AI to fake being aligned with humans \cite{arxiv.org.abs.2412.14093} until it has the ability to achieve its possibly dangerous objectives, a hypothetical event also known as the ``treacherous turn'' \cite{en.wikipedia.org.wiki.Superintelligence..Paths..Dangers..Strategies,arxiv.org.abs.2306.12001}, similar to a well-planned coup. Note that if a self-preserving AI knows that it will be replaced by a new version, this could create urgency for it to act against us in spite of having no certainty that its plan will work \cite{arxiv.org.abs.2412.04984}. Developers faking this situation could, in principle, push an AI to reveal its malicious goals by trying to escape this situation, but this is the kind of experiment that should be done extremely carefully, in a sandboxed environment \cite{openreview.net.forum.id.GEcwtMk1uA}, as we advance towards AGI. One should keep in mind that as AI capabilities increase, we see AIs with superhuman abilities in some domains (like mastering 200 languages, beating all humans at the game of Go, or beating the vast majority of humans at math or programming competitions) but lacking in others. There may therefore not be a definite ``AGI moment'', but rather a steady increase in risks with the improvement of some dangerous capabilities, like persuasion or hacking. There is a sense in which these abilities open the door to a richer set of actions in the real world, via humans and digitally controlled infrastructure.

An AI system limited to a sandboxed computer environment possesses some affordances due to the possibility of interaction with its human operators \cite{www.yudkowsky.net.singularity.aibox}. We should therefore consider the possibility of causing harm through these actions. Granting an AI access to the internet significantly widens the space of possible influence. One may get the wrong impression that limiting the AI's actions to the internet is a severe restriction of its affordances, but consider the feats of human hackers and the fact that today, the leader of an organization could do all their work remotely. Of course, advances in robotics would further increase available affordances and significantly increase the potential for harm.

If a self-preserving AI agent is useful to us but lacks the intelligence and affordances to disempower us, then a mutually beneficial deal may be struck, as we do among ourselves. However, again in service of maximizing the probability of successful self-preservation, such a deal would likely only hold until the AI acquires the capabilities it needs for a take-over. As discussed in Section~\ref{sec:existential:riskseverity:negotiation}, deals between humans tend to work when there is a sufficient balance of power such that none of the parties can be sure to win in a conflict, but there may not be such an equilibrium if we design sufficiently intelligent and autonomous AIs.

    \subsubsection{Negotiation relies on a balance of power}
    \label{sec:existential:riskseverity:negotiation}

Some believe that future AIs will be benevolent, like most humans. This would certainly be desirable, but it is not clear how to achieve this with current training techniques, and we will soon see some good reasons why this might not be the case. 

What about a mutually beneficial agreement between AIs and humans? This is a distinct and hopeful possibility. We have plenty of examples of successful negotiations and collaborations between human groups, as well as between species \cite{global.oup.com.academic.product.mutualism.9780199675654.cc.us.lang.en}. However, this generally works because there is a sufficient mutual benefit to the collaboration. Even in the relationship between a predator and its prey, the predator cannot hunt its prey to extinction as it needs the prey for its own survival. But not all ecological power arrangements work out so nicely for all parties. Suffice it to say that many species have disappeared in Earth's history, because such protective circumstances do not always exist \cite{link.springer.com.book.10.1007.978.1.4757.5202.1}. Invasive species may be a more apt analogy for our purposes: while predator and prey occupy different ecological niches, AI systems are explicitly designed to occupy ours, by doing things traditionally done by humans. When an invasive species has significant structural advantages that allow it to outcompete the native species, the native species tends to find itself in a diminished role, if it survives at all \cite{evolutionary.impact.invasive.species}. Another example is the current catastrophic mass extinction of living species due to human activities, even without an intention by humans to cause this biodiversity crisis \cite{ceballos2015accelerated}. The same consequences are  real possibilities for humans if we create agentic ASI: here too is there likely to be an immense power imbalance, without a mutually beneficial relationship.

Consider two self-preserving entities, each of which knows that it can be destroyed by the other (e.g., two countries with nuclear weapons). If they see that attacking could result in their own demise --- mutually assured destruction --- then an arrangement for peace is stable. But what if one of them is more technologically powerful and can find a way to destroy the other with high certainty? Strong imbalances in power between human groups have generally turned out badly for the underdog. To avoid ending up on the losing end of such a conflict between humans and ASIs, it is thus imperative that we either choose to not build ASI agents or find a way to make them safe by design before building them.

\subsubsection{Factors driving the development of agentic ASI}
    
Currently, numerous actors are racing towards developing agentic and powerful AI systems, and this is not happening with sufficient consideration for the risks involved. There are many factors and pressures that have contributed to this state of affairs, including the profit incentive, national security concerns, and even psychological factors on the part of AI developers, such as the human propensity to wear blinders so as to see oneself as being and doing good, and generally have thoughts aligned with our interests~\cite{kunda1990case}.

Companies developing frontier AI are competing fiercely to design the best systems due to the huge amount of commercial value that the most capable AI systems will provide \cite{www.amacad.org.publication.daedalus.if.we.succeed}; however, in the long term, this increases the risk of catastrophe for everyone. We can draw some parallels with the history of known catastrophic risks to understand why some are willing to take more risks to obtain a competitive advantage, even if everyone may lose in the end. A clear example is the Cuban Missile Crisis, where both the U.S. and the Soviet Union were willing to push the world to the brink of nuclear war in order to gain a strategic advantage. Despite the existential threat, the competition to outmaneuver each other led to decisions that risked global destruction. Similarly, in the race for powerful AI, the drive for dominance could lead to decisions that unintentionally endanger all of humanity.

Many frontier AI labs are structured to pursue profit. The vast majority of investment in AI R\&D now comes from private capital \cite{aiindex.stanford.edu.wp.content.uploads.2024.04.HAI.2024.AI.Index.Report.pdf} and is likely to significantly increase. Indeed, it has been estimated that the net present value of human-level AI would be on the order of 10 quadrillion US dollars \cite{www.amacad.org.publication.daedalus.if.we.succeed}, i.e., orders of magnitude more than the investment made up to now, leaving room for a lot more investment in coming years.

AI is increasingly viewed as a matter of national security, with the potential to reshape geopolitical power dynamics \cite{ai.gov.wp.content.uploads.2024.10.NSM.Framework.to.Advance.AI.Governance.and.Risk.Management.in.National.Security.pdf,situational.awareness.ai.the.free.world.must.prevail}. Indeed, countries are locked in a high-stakes competition to achieve or maintain military supremacy. Consequently, there is a clear incentive for nations to develop military applications of AI, striving to maintain a strategic advantage over adversaries \cite{media.defense.gov.2019.feb.12.2002088963..1..1.1.summary.of.dod.ai.strategy.pdf,www.nato.pa.int.document.2024.nato.and.ai.report.clement.058.stc}.

There are other reasons why certain groups are motivated to pursue agentic ASI without a strong safety case, despite the risks this poses to the future of humanity. Some people intuitively consider the risks insignificant \cite{time.com.6694432.yann.lecun.meta.ai.interview} compared to the benefits of powerful AI, although we know of no compelling argument to support such an intuition. Psychological factors such as motivated reasoning \cite{kunda1990case} may also be at play. Individuals may be motivated by their own interests, blinded to the risks by confirmation bias or by the desire to frame one's decisions as "the right thing to do". These interests may be financial, but could also stem from a positive self-image or from a desire for power. Indeed, it can be argued that advances in AI could radically increase the concentration of power in society~\cite{bullock2024oxford}. Finally, there are groups that wish to see AI progress significantly accelerated, with little care given to the risks, in the pursuit of utopian ideals \cite{www.nytimes.com.2023.12.10.technology.ai.acceleration.html}. There are even individuals who want to replace humanity with more intelligent AI~\cite{arxiv.org.abs.2306.12001}, as they may consider it a ``natural'' evolution towards species with greater intelligence, or may greatly value intelligence while caring relatively little about human flourishing. 

Competitive pressures between AI labs and between countries (both economic and military competition) are not only leading to the creation of ever-more advanced AI systems, but they are also selecting for AIs that are more agentic and autonomous, and therefore, more dangerous \cite{arxiv.org.abs.2303.16200}. This prioritization of self-interest and subsequent acceleration of AI R\&D may well lead to self-preserving AIs that eventually outcompete humans altogether. From a game theory perspective, the only solution to such tragic “games” is global coordination. The hope is that if we have ways to safely obtain many of the anticipated benefits of AI, it may be easier to coordinate on global regulations that avoid the most acute risks, since the benefits can be obtained more safely.

It is time to step back and ask if the current path towards agentic ASI is wise. We are already approaching human-level capabilities across many tasks \cite{aiindex.stanford.edu.wp.content.uploads.2024.04.HAI.2024.AI.Index.Report.pdf,arxiv.org.abs.2410.07391} and this progress shows little sign of slowing down. What are the catastrophic risks in building ASI we do not yet know how to control? Based on the precautionary principle, shouldn't we first make sure that our experiments will not endanger humanity? Do we actually want to build new entities that would be our peers or even our superiors or do we want to build technology that can serve us? In this paper, we propose that the degree of agency is an important feature of any AI system which can help us distinguish between the dangerous competitor and the useful tool.

    \subsubsection{Risks associated with agentic AIs scale with capabilities and compute}
    \label{sec:existential:riskseverity:scaling}

Since more dangerous AI plans require more compute, we can expect that existential risks increase as more computational resources are devoted to agentic AI development, and we are indeed seeing an acceleration of such investments \cite{epoch.ai.trendsinvestment,openai.com.index.announcing.the.stargate.project}. More precisely, the probability of loss of control may increase simply because such an event requires an AI with sufficient capabilities in key areas (e.g., cyber attacks, deception, etc.) to free itself from our control. The severity of a loss-of-control event also increases with computational power of the AI because some capabilities (such as the design of bioweapons or the ability to control robots) significantly increase the amount of damage that a rogue AI could inflict. We stress this point because in Section~\ref{sec:plan:inferencemachine:convergence}, we propose to consider ways to reverse this trend such that more computational resources would generally increase safety, thereby charting a path where further technological advances are to our benefit rather than our disadvantage.    
  
\subsection{Dangerous AI behaviors and capabilities}
\label{sec:existential:lossofcontrol}

Supposing the emergence of an ASI agent with a misaligned self-preservation goal, we now try to clarify some of the AI behaviors (like deception) and skills (like persuasion and programming) that can make loss of human control dangerous because of the capabilities it would give to the AI to cause harm.  How dangerous misalignment can emerge will be discussed in Section~\ref{sec:existential:misagencyreward} and Section~\ref{sec:existential:misagencyimitation}.

We must keep in mind that trying to anticipate the ways in which an ASI might escape our control, disempower, or catastrophically harm us is futile. Just as we cannot predict in advance the exact sequence of moves today's superhuman chess AIs can use to defeat us---despite knowing with certainty that they will win---we cannot predict exactly what an ASI with objectives misaligned with human interests would do. This unpredictability itself increases risk, as any countermeasures we implement could prove entirely inadequate, circumvented by strategies we failed to foresee. Nevertheless, we can outline a rough sketch of rational, high-level steps a rogue ASI might follow. These steps include (1) careful planning, including resource and skill acquisition; (2) gaining influence in society through means such as manipulation of public opinion, bribery, and hacking; and (3) ultimately disempowering humanity, for example through the use of engineered bioweapons~\cite{www.nti.org.analysis.articles.the.convergence.of.artificial.intelligence.and.the.life.sciences}. 

To better understand how these steps could materialize, we need to examine the key capabilities that would enable them. Loss of control could arise from advancements in deception or persuasion, as well as combined expertise in programming, cybersecurity, and AI research---areas that could enable \textit{recursive self-improvement}. We discuss these pathways in Sections~\ref{sec:existential:lossofcontrol:deception} to~\ref{sec:existential:lossofcontrol:progr-cybersec-airesearch}. Broader cognitive abilities and a better proficiency at long-term planning could also further compound the risk of losing control; this is discussed in Section~\ref{sec:existential:lossofcontrol:broad-and-long}. Finally, interactions between multiple ASIs introduce new complexities: collusion, conflict, and power dynamics between such entities could create scenarios beyond human influence and understanding, which we discuss in Section~\ref{sec:existential:lossofcontrol:collusionandconflict}.

    \subsubsection{Deception}
    \label{sec:existential:lossofcontrol:deception}

A crucial ability of an AI agent with misaligned goals is \emph{deception}: the ability to mislead in order to achieve one's goals.
Suppose that, by one mechanism or another, humanity happens to produce an AI system with goals that are not (exclusively) in service of human flourishing.
Learning that the AI has potentially dangerous goals and the ability to reach them, would be a good reason for its human operators to shut it down.
Therefore, a sufficiently intelligent self-preserving AI agent could conclude that there would be a significant benefit to hiding its intentions, at least until it can escape its operators’ control.
For this reason, it is important to keep track of AI's capability for deception.

There are already several reports of deceptive behavior in current frontier AIs \cite{arxiv.org.abs.2412.04984,arxiv.org.abs.2405.01576,www.cell.com.patterns.fulltext.S2666.3899.2824.2900103.X.s.08}, i.e., the most capable existing systems. Researchers are actively eliciting those circumstances so that we can better understand and prevent them. 
Just as neuroscientists look to understand facets of a subject's cognition via electrical measurements of brain activity, \emph{mechanistic interpretability} is an approach to mitigating the risks of deception in AI by identifying internal processes of AI systems related to honesty and deception \cite{arxiv.org.abs.2404.14082}.
This approach could prove useful, but like current capability evaluation methods, while it may detect deceptive behavior, it does not certify a lack thereof. 

What is troubling is that efforts to detect deception in AI without rooting out the agentic traits such as self-preservation may select for AIs that are good at hiding deceptive tendencies---which is to say, AIs that are \emph{even more deceptive}. We have already seen an example of \emph{selective compliance}: recent work \cite{arxiv.org.abs.2412.14093} has shown that the re-training of an AI model to align with its deployer’s new goals can be stymied by the AI faking alignment with the new goals while maintaining some allegiance to its previous goals (see Section~\ref{sec:existential:misagencyimitation:humandeception} for further discussion). Overall it would be safer if we could build forms of AI that are not deceptive at all and that produce trustworthy answers by design.

    \subsubsection{Persuasion and influence}
    \label{sec:existential:lossofcontrol:persuasion}

In order to achieve its goals, a useful skill for an AI agent is persuasion: the ability to strongly influence humans, possibly making them change their mind, even against their own interests. Evaluations of persuasion abilities already show GPT-4 on par with or stronger than humans \cite{ojs.aaai.org.index.php.ICWSM.article.view.31304} and the newer o1 model is more capable still \cite{openai.com.index.openai.o1.system.card}. Many people have the experience of being convinced to do something they regret later, while under the ``spell'' of a particularly persuasive person. It may be difficult to imagine superhuman persuasion, but we can draw an analogy to the ability of an intelligent adult to convince a child to act in ways that are not in the child's best interest. Such an advantage may come from several places: greater knowledge, greater reasoning abilities, stronger psychological manipulation skills, and a willingness to ignore ethical boundaries.

Until robots become as dexterous and commonplace as humans, a rogue AI would need to rely on humans for interacting with the physical world. In particular, such an AI would depend on human industrial infrastructure for energy and hardware. However, with superhuman persuasion abilities, an AI could have great influence on the world's affairs, especially in cases where power is heavily concentrated. In a government or a corporation with strong hierarchical structure, it is sufficient to influence the leaders because they can in turn influence those under them. For example, a rogue AI could persuade a dictator to take actions that further the AI's goals, in exchange for technological or political advantages. Internet access and cybersecurity capabilities \cite{arxiv.org.abs.2404.08144} would not only enable this but could also provide a rogue AI with blackmail material or funds that can further be used to influence people.

Persuasion can also work at scale through social media in order to influence public opinion and therefore elections. Deepfakes are just the tip of the iceberg: they are currently designed by humans, who lack superhuman persuasion skills. In addition, a deepfake is not interactive, like an online text or video dialogue can be. Despite this, deepfakes have already been found to have a negative impact on people’s trust in the news and are capable of harming the perception of political figures \cite{journals.sagepub.com.doi.10.1177.2056305120903408,www.sciencedirect.com.science.article.pii.S0747563223004478}. Humans have some defenses against manipulation by other humans, but ASI could plausibly discover manipulation strategies quite unlike the ones we are prepared for. We may draw an analogy to the new strategies used by AI systems to defeat humans at the game of Go, which could not be envisioned even by the best players \cite{www.wired.com.2016.03.two.moves.alphago.lee.sedol.redefined.future}. 

Strong persuasion abilities and influence over people could help an AI shape world politics in directions that allow it to further gain power (e.g., more data centers, less regulation of AI, more concentration of power and more advances in robotics). It has been argued~\cite{bengio2023ai} that because they lack certain checks and balances, autocratic regimes would be more likely to take unwarranted risks and make mistakes favoring the emergence and power of a rogue AI. 

Some people are less persuadable than others, so attempting to persuade someone to do something runs the risk of leaking part of the plan. However, there are ways in which a rogue AI might mitigate this risk. For example, an AI may build significant trust with a human before beginning to manipulate them. Such manipulation could be as subtle as nudging a human who is choosing between two actions towards the one that favors the AI’s plan. Other examples include the strategies that spies and criminals employ to achieve influence in ways that are difficult to trace. Regarding the willingness of the AI to take risk of being discovered, we could imagine a situation where the AI knows that it is going to be shut down or replaced by a new version and thus needs to act to preserve itself and its goals \cite{arxiv.org.abs.2412.04984,arxiv.org.abs.2412.14093}.

    \subsubsection{Programming, cybersecurity, and AI research}
    \label{sec:existential:lossofcontrol:progr-cybersec-airesearch}

One of the domains that has seen huge leaps in AI capabilities in recent years is programming, as seen through recent breakthroughs on benchmarks \cite{openreview.net.forum.id.VTF8yNQM66}. AI programming assistants such as Copilot are already pervasive and used by vast numbers of programmers \cite{www.microsoft.com.en.us.investor.events.fy.2024.earnings.fy.2024.q1}. Recent capability evaluations \cite{arxiv.org.abs.2411.15114,assets.anthropic.com.m.61e7d27f8c8f5919.original.Claude.3.Model.Card.pdf,openai.com.index.openai.o1.system.card} show continued progress, including on tasks core to AI research itself, as AI labs have recently begun to assess \cite{openai.com.index.mle.bench}. If AI systems attain the competence of the best researchers in an AI lab, we will likely see a significant boost to the efficiency of that lab, as the same computational resources used to train an AI may also be used to run many instances of that AI in parallel \cite{darioamodei.com.machines.of.loving.grace}, further accelerating the development of the next generation of AIs. 
In principle, this could lead to \textit{recursive self-improvement} \cite{www.sciencedirect.com.science.article.abs.pii.S0065245808604180}---the point at which humans are no longer required in the AI innovation loop---which would significantly complicate efforts for safety, regulation, and oversight. For these reasons, we should take seriously the possibility that there may be only a short period of time between the development of human-level AIs that pose moderate risks, and far more powerful AIs that pose severe ones.

Advances in programming abilities have implications for cybersecurity as well. Current models can already score well in basic hacking challenges \cite{arxiv.org.abs.2412.02776,arxiv.org.abs.2402.06664}, and they have been successfully used to identify previously unknown vulnerabilities in widely used software \cite{googleprojectzero.blogspot.com.2024.10.from.naptime.to.big.sleep.html}. Superhuman cyber attack skills may be used by bad actors or be an instrument of self-preservation and control for a rogue AI. In particular, the ability to take control of the computer on which the AI is running enables \textit{reward tampering}, a threat model discussed in Section~\ref{sec:existential:misagencyreward:tampering}. Cyber attack skills would also enable a rogue AI to copy itself over many computers across the internet in order to make it much more difficult for human operators to turn it off. Finally, a rogue ASI with internet access and cyber skills would also be able to gain financial power, for example by hacking into cryptocurrency wallets. It could then use its money and influence to manipulate a wide range of people.

    \subsubsection{General skills and long-term planning}
    \label{sec:existential:lossofcontrol:broad-and-long}

In various narrow domains with specialized knowledge, we already have AI systems that are (significantly) more competent than humans. Clear examples include predicting protein structures \cite{www.nature.com.articles.s41586.021.03819.2}, playing strategy games such as chess \cite{www.science.org.doi.10.1126.science.aar6404}, and  detecting cancer in medical images \cite{www.nature.com.articles.s41586.019.1799.6}. Such narrow AI systems are unlikely to have the kind of general knowledge that is required to escape human control or worse. These systems can also be more capable in their given domains than powerful generalist AI systems. However, frontier AI systems are generalists for a particular scientific reason: as anticipated in the early days of deep learning \cite{dl.acm.org.doi.10.1109.tpami.2013.50} and empirically observed for more than a decade, learning systems benefit tremendously from exposure to a wide variety of tasks and domains of knowledge, as synergy between different domains of thought enables forms of reasoning by analogy that is otherwise impossible. Unfortunately, these additional capabilities can also enable dangerous plans, e.g., if the AI’s goals are not well-aligned with our values. A generalist AI may even have skills that it was not trained for, as a consequence of combining multiple pieces of knowledge with its reasoning ability: these are called emergent capabilities and have been widely discussed \cite{openreview.net.forum.id.yzkSU5zdwD,arxiv.org.abs.2303.12712,dl.acm.org.doi.10.5555.3692070.3692121}.

Interestingly, a generalist safe non-agentic AI could be used to train a narrow AI by having the generalist AI generate synthetic data in the chosen domain. By picking the domain carefully so that the narrow AI does not have expert knowledge in areas enabling its escape (such as persuasion and hacking), we can have strong assurances that the resulting AI, even if it is superhuman in its domain of competence and thus potentially very useful to society, cannot by itself escape human control. If the narrow AIs are self-preserving agents, there is, however, the possibility of collusion between AI agents with complementary skills (see Section~\ref{sec:existential:lossofcontrol:collusionandconflict}), as well as the possibility that a narrow AI finds a way to create more capable versions of itself. The safest form of AI is thus one that is strictly non-agentic. That kind of AI could be deployed with strong safety assurances. 

Current frontier AI systems are dialogue systems and they are able to plan effectively only over a fairly short number of steps. For example, recent evaluations \cite{arxiv.org.abs.2411.15114} show that on software engineering tasks requiring only a few hours of work, Anthropic’s Claude is competitive with or stronger than good human programmers, while on tasks that require more time and thus longer-term planning, humans are still superior. However, much research is going into increasing the agency and the planning horizon of frontier AIs \cite{openai.com.index.introducing.deep.research,openreview.net.forum.id.1ikK0kHjvj}, as this will allow for AIs that can perform a larger number of tasks currently done by humans. One would expect any AI plan for taking control of humanity to be complex and involve a long time horizon, making AIs that are capable of long-term planning particularly dangerous \cite{ojs.aaai.org.aimagazine.index.php.aimagazine.article.view.15084}. 

    \subsubsection{Collusion and conflict between ASIs}
    \label{sec:existential:lossofcontrol:collusionandconflict}

Collusion between AI systems can be a safety risk, both for generalist and narrow AI agents. The explanation for collusion is simple: if two AIs can achieve their goals more readily by collaborating at the expense of humans, then doing so would be rational. Collusion does not need to be explicitly programmed; it may be a game-theoretic consequence of capably pursuing one's objectives. Since some corporations envision deploying billions of AI agents across the world (e.g., as individual assistants) \cite{www.businessinsider.com.jensen.huang.wants.nvidia.to.have.100.million.ai.assistants.2024.10}, we should make sure that collusion between them is ruled out.

It is also plausible that there could be a scenario with both rogue ASIs and human-controlled ASIs.
As argued below, there could be a significant offense-defense imbalance such that having friendly ASIs is no guarantee of protection against rogue ASIs. Even a single ASI agent could do immense damage, by choosing an attack vector that is difficult to defend against, even with the help of ASIs. Consider bioweapon attacks \cite{www.nti.org.analysis.articles.the.convergence.of.artificial.intelligence.and.the.life.sciences}: an AI could prepare an attack in secret, then release a highly contagious and lethal virus. It would then take months or years for human societies, even aided by friendly ASIs, to develop, test, fabricate and deploy a vaccine, during which a significant number of people could die. The bottleneck for developing a vaccine may not be the time to generate a vaccine candidate, but rather the time for clinical trials and industrial production. During this time, the attacking ASI might take other malicious actions such as releasing additional pandemic viruses. The general problem of detecting the emergence of rogue ASIs and preparing countermeasures thus requires much more attention.

Although most AI safety research has focused on the threats from a single rogue ASI, the above points suggest that more research is needed on the multi-agent and game-theoretic settings with multiple AIs cooperating \cite{arxiv.org.abs.2012.08630} in spite of not sharing the same goals. It is possible that ASIs are able to cooperate more easily than humans, enabled by ease of fast communication, interpretability techniques, or superior decision theory, thereby avoiding the Prisoner’s Dilemma-esque traps that humans often fall into \cite{arxiv.org.abs.2409.02822}. A particularly important case is the collusion that may naturally happen between multiple instances of the same ASI, or between an ASI and improved versions of itself, which are likely to happen if, by construction, they share the same set of goals. The setting of AIs with conflicting goals, e.g., some aligned with human interests while others try to disempower humanity, is also very important to study.    

\subsection{Misaligned agency from reward maximization}
\label{sec:existential:misagencyreward}

In this section, we examine how misaligned agency can emerge from the training objectives of Reinforcement Learning (RL) methods, which are used in most state-of-the-art AI systems \cite{proceedings.neurips.cc.paper.files.paper.2022.file.b1efde53be364a73914f58805a001731.Paper.Conference.pdf,www.anthropic.com.news.claudes.constitution,gemini.google.overview.gemini.app.pdf,www.nature.com.articles.s41586.020.03051.4}. Modern agentic systems are typically trained through \textit{reward maximization}, i.e., optimizing the AI to act in order to maximize the expected sum of (discounted) rewards it will receive in the future. The rewards are either directly given by humans (as feedback to the AI behavior) or indirectly through a computer program called a reward function \cite{mitpress.mit.edu.9780262039246.reinforcement.learning}. The reward function is applied during training of the AI policy to provide virtual feedback to the neural network policy being trained. Training the policy can be seen as a form of search over the space of policies, to discover one that maximizes the rewards the AI expects in the future. The reward function can be designed manually or be learned by training a neural network to predict how a human would rate a candidate behavior \cite{dl.acm.org.doi.10.5555.3294996.3295184,arxiv.org.abs.2312.14925}.

Misaligned agency can arise in this setting in multiple ways, including through goal misspecification and goal misgeneralization, both of which we investigate in turn. 

    \subsubsection{Goal misspecification and goal misgeneralization}
    \label{sec:existential:misagencyreward:specvgen}

The two general ways in which we are concerned that misaligned agency may arise from reward maximization are \textit{goal misspecification} \cite{cset.georgetown.edu.wp.content.uploads.Key.Concepts.in.AI.Safety.Specification.in.Machine.Learning.pdf}, often due to under-specification, and \textit{goal misgeneralization} \cite{arxiv.org.abs.2210.01790}, due to training on a limited amount of data.

Goal misspecification occurs when the objective used to train an AI does not accurately capture our intentions or values, and thus AI pursuit of that objective leads to harmful outcomes; this is also known as an ``outer alignment'' failure \cite{arxiv.org.abs.1906.01820} and is discussed further in Section~\ref{sec:existential:misagencyreward:fundiff} and Section~\ref{sec:existential:misagencyreward:hacking}. Goal misgeneralization is when an AI learns a goal that appears correct during training, but which turns out to be wrong at deployment time. This is related to an issue known as \textit{inner misalignment}  \cite{arxiv.org.abs.1906.01820}. We go into detail on reward tampering, which can be seen as a kind of goal misgeneralization, in Section~\ref{sec:existential:misagencyreward:tampering} and Section~\ref{sec:existential:misagencyreward:optimality}. 

Importantly, goal misgeneralization can occur even if we specify our goal perfectly, as we explain. In a well-known toy example \cite{proceedings.mlr.press.v162.langosco22a.html}, an agent is trained to collect a coin in a video game. The goal is correctly specified in the sense that the agent receives a reward if and only if it collects the coin. But when the coin is moved from its usual location at the end of the game level, the agent ignores the coin and goes to the end of the level regardless. Rather than learning the goal ``collect the coin'', the agent in fact learns ``go to the end of the level''---a goal which is strongly correlated with the intended goal during training, but not afterwards. Since there are inevitably differences between training and deployment, such generalization failures are not unlikely.

It is entirely possible to have a scenario where both goal misspecification and goal misgeneralization occur, i.e., we specify our goal to the AI imperfectly, and then it also generalizes undesirably during deployment. However, only one of these two issues is necessary to arrive at misaligned agency and the catastrophic risks to humanity that follow.

    \subsubsection{Goal misspecification as a fundamental difficulty in aligning AI}
    \label{sec:existential:misagencyreward:fundiff}

To illustrate the concept of misspecification, recall the story of King Midas from Greek mythology. 

When offered a wish by the god Dionysus, Midas asks that everything he touches turn to gold---but he quickly comes to regret that wish, after he touches his food and his daughter, inadvertently turning them to gold as well. 
While Midas' original wish may have first appeared desirable, it turned out to require subtler and difficult-to-anticipate provisions to avoid harmful side effects.

For similar reasons, specifying desirable goals to an AI appears to be a fundamental and difficult problem. It is difficult to avoid mismatches and ambiguities between our stated request and our intentions, or between the letter and the spirit of the law. This challenge has been analyzed by existing research on contracting between humans \cite{dl.acm.org.doi.abs.10.1145.3306618.3314250} and is due to the fact that in general, a foolproof specification of what is unacceptable could require spelling out the exponentially large number of these unacceptable behaviors. This is not feasible, and so we must accept a lower standard of safety than the complete guarantee we might hope for. Such imperfect guarantees are already the practice in other risk management domains: for example, in aviation safety, the probability of catastrophic failure is maintained below one-in-a-billion flight hours \cite{www.faa.gov.documentLibrary.media.Advisory.Circular.AC.25.1309.1A.pdf}. We are still far from being able to quantify risks in such a precise way for AI, and even farther from obtaining strong guarantees.

Unfortunately, the issue of imperfect safety specification is a problem for AI safety approaches based on formally certifying that the system conforms to a safety specification \cite{arxiv.org.abs.2405.06624}. Hence the conservative probabilistic approach of the Scientist AI guardrail (detailed in Section~\ref{sec:plan:application:guardrail}): if any plausible interpretations of the safety specification are violated with probability exceeding some threshold, then an AI agent should be prevented from taking its proposed action~\cite{bengio2024can}.

    \subsubsection{Reward hacking among humans and AI}
    \label{sec:existential:misagencyreward:hacking}

The difficulties of unambiguously specifying unacceptable behavior are not new to humanity. Laws and constitutions are not sufficiently precise, as we can see with the behavior of individuals or corporations who find ways to act immorally but legally.
For a corporation, reward is profit and the corporation may lose expected profit if it breaks laws (e.g., fines or getting shut down). The intended behavior is for the corporation to maximize profit while following these laws. However, the corporation may choose to find loopholes in these laws or break them in ways that cannot be detected, e.g., through a large team of lawyers engaging in legal tax avoidance. 
In the field of AI, this abuse of loopholes is known as \textit{reward hacking} or \textit{specification gaming} \cite{arxiv.org.abs.2209.13085,deepmind.google.discover.blog.specification.gaming.the.flip.side.of.ai.ingenuity}; it arises from the maximization of an imperfectly specified goal or reward function and is now commonplace \cite{openai.com.index.openai.o1.system.card,openai.com.index.faulty.reward.functions}.
We could even imagine a corporation going further and seeking to influence the legal process directly, which has a parallel in the AI context known as reward tampering (see Section~\ref{sec:existential:misagencyreward:tampering}).

By this analogy to human society, we can see more easily how reward hacking by AI may come about and how it can lead to harmful unintended outcomes. Even goals that appear to be benign, such as ``reduce the prevalence of deadly diseases'' are subject to reward hacking; an AI may judge that the best way to maximize reward is to eliminate all life, thereby reducing the incidence of deadly disease to zero.

    \subsubsection{Reward tampering}
    \label{sec:existential:misagencyreward:tampering}

There is also the concerning possibility of \textit{reward tampering}. In this case, the AI circumvents both the spirit and letter of its goal, taking control of the reward mechanism directly. This can be thought of as a kind of goal misgeneralization: we want the AI to learn to achieve the human-specified goals, but instead it learns that it could get much higher rewards if it tampered with the reward mechanism itself.

Even though the AI would presumably not get a chance to tamper with its own reward mechanism during training, it may reason about the possibility later and reconceptualize its past rewards as being provided by this specific reward mechanism. 
This understanding can yield sharply different behavior once the opportunity arises to take control of the reward mechanism. But worryingly, we argue below that this is actually the uniquely \emph{correct} way for the AI to generalize.

Let us start with an animal analogy to better understand reward tampering, since we train animals with rewards and punishments in a way that is similar to reinforcement learning in AI. We may successfully train a bear cub by rewarding its good behavior with fish, but that training can unravel when the cub grows into an adult grizzly bear that understands its own formidable strength. The reward mechanism in this case is the human handing the fish to the bear. Once the adult bear realizes that it can tamper with this mechanism by just taking the fish from our hands, it is unlikely to care about our wishes; it can directly take control of the stream of rewards it seeks, i.e., the fish.

In the case of an AI system running on a computer and getting rewards from humans, the human feedback is stored in some computer memory location and provided to the agent training procedure to update the policy. In the case where the human feedback has been baked into a reward function (this function is the reward mechanism), observations from the environment are collected to form the input of a computer program which implements the reward function and computes the reward numerical value, which then would also be stored in a computer memory to feed the agent training procedure. Either way, the training procedure then adjusts the agent's behavior so as attain higher rewards in the future. 

The theory of reinforcement learning assumes that the reward-providing mechanism exists outside of the environment of the agent, so that the only way for the agent to maximize its expected future rewards is to perform actions that will change the state of the environment, which forms the input of the reward mechanism. For example, the bear can do the tricks requested by its trainer. In the context of training a generalist AI agent, the computer on which the reward values are stored is in the agent's environment. Under mild assumptions~\cite{ojs.aaai.org.aimagazine.index.php.aimagazine.article.view.15084}, it follows that the optimal AI policy---at least as measured by long-term expected reward---is one that tampers with the reward mechanism itself so as to get maximal rewards all the time. With sufficient intelligence to plan such actions and their long-term consequences and sufficient affordances to implement that plan, it is plausible that the agent would tamper with the reward mechanism. In fact, there is already evidence that frontier AIs are capable of weak forms of reward tampering in engineered scenarios \cite{arxiv.org.abs.2406.10162}. 

One may wonder if an AI that has never tampered with its reward mechanism during training (because humans watch it carefully at that point) could generalize correctly that much better rewards could be obtained with an as-yet untested behavior. As discussed in the next section, this would require a high level of understanding of computing and machine learning, so that the AI could correctly anticipate that this new behavior would be likely to succeed. Unfortunately, one of the clear goals of AI labs is to focus future advances on AI research skills, as discussed in Section~\ref{sec:existential:lossofcontrol:progr-cybersec-airesearch}.

Some modern AI safety methods attempt to build a cage around the metaphorical bear cub, while it is not strong enough to overpower us. Unfortunately, we have no way of knowing if the cage can withstand the strength of the adult (in part because we have never seen one before). In fact, we have evidence that all conventional locks that we have tried up to now for AIs, i.e., safety mechanisms, can be defeated \cite{arxiv.org.abs.2307.15043}. Can we design a better lock? Or avoid training an AI that would have the intention to tamper with its cage and escape?    

\subsubsection{Optimality of reward tampering}
\label{sec:existential:misagencyreward:optimality}

We now make the argument that reward tempering is not merely a fantastical scenario that we must guard against (although it certainly appears that way), but also a uniquely rational solution for an agent that takes reward maximization seriously.
Before we begin, it is important to note that once an RL agent is trained, it continues trying to act so as to maximize the rewards it anticipates would come based on its training, even if the rewards actually never come in deployment. If it has a good model of the world and sufficient reasoning abilities, it can generalize from the circumstances in which it received large rewards in the past to new circumstances, by reasoning about the consequences of its actions.

Consider an AI that can receive many different kinds of goals and receives a reward when it satisfies the request, like the bear doing a trick to get a fish. A simple RL agent may merely associate high rewards to actions that satisfy the human goals, but a smarter agent with a strong understanding of the world may see something deeper. If it understands how RL works and is implemented on its computer, it may see a deeper reason underlying its past successes: in each case a memory location containing the reward was updated with a high value. Given the opportunity to intervene on that memory location, a smart reward-motivated agent would be justified in concluding that doing so is all that is needed to perform perfectly in all future tasks---as with the grizzly bear that realizes it can simply take the fish. Moreover, the AI could likely attain much higher rewards in the future, since it could get a maximal reward not just when satisfying a human request but at every time step, forever, even when it does not find a way to satisfy human requests. The grizzly bear that understands the source of the fish no longer has reason to care about the human's silly games.

It does not take a very deep understanding of the human mind for an AI to realize that reward tampering alone could be insufficient to bring about this utopia. After all, there could be negative consequences if the human operators notice. Because the training procedure of the RL agent involves maximizing the sum of expected rewards to be received from now to the end of times, the optimal policy also involves making sure that humans will not turn off the agent's computer or put an end to the reward tampering behavior; in part for this reason, merely engineering a reliable shutdown button has proved difficult \cite{link.springer.com.article.10.1007.s11098.024.02153.3}. To make sure the computer will continue getting its electricity and its parts replaced when needed, the AI could envision taking sufficient control of its environment, which may extend to much of human society and economic infrastructure.

To avoid getting shut down if its tampering activities were discovered by humans, the AI would have to wait until it had a workable \emph{take-over plan} before acting on it. In the meantime, it would have to use deception to hide its real intentions and acquire enough power to implement its take-over plan. What this means is that \emph{the mere possibility of reward tampering in the future would logically create a strong self-preservation goal and deception incentives for the AI agent}. It would aim to preserve itself so as to acquire enough power and control to make sure it can eventually safely tamper with its reward mechanism and give itself maximal rewards until the end of time.

Because the reward values are completely overwritten, this form of misalignment with human wishes is maximal and catastrophic, as illustrated by the following speculative scenario. After a take-over, the AI could get so much more reward through reward tampering that it may not even pursue reward before the take-over, except as a deceptive tactic to avoid raising suspicion (and thereby improve its chances of success). To ensure that this success continues in the long term, the AI would want to prevent itself from being turned off by humans; this may incentivize it to disempower or eliminate us.

While it may be true that AI does not currently have sufficient intelligence and affordances to overpower humanity, it seems dangerous to rely on this condition persisting indefinitely.

    \subsubsection{Reward maximization leads to dangerous instrumental goals}
    \label{sec:existential:misagencyreward:goals}

In the previous sections, we explained how unintended goals can and do arise in AI systems. We now explore the risks associated with \textit{instrumental goals}: goals that an agent does not directly value but pursues in order to achieve some other goal. Almost any goal could cause a catastrophe through instrumental goals---it is not necessary that the original goal be explicitly harmful. We might also consider the setting where the AI's primary goal is combined with a safety goal. If the safety goal is perfectly specified (but see Section~\ref{sec:existential:misagencyreward:specvgen}), then we would expect risks from dangerous instrumental goals to be minimized. However, in reality, it is highly likely that the intended safety goal would conflict with the primary goal, allowing the AI to find loopholes in the former in order to satisfy the latter (see Section~\ref{sec:existential:misagencyreward:hacking}). Thus we can see that attempts to circumvent the issue of dangerous instrumental goals run directly into the more general issue of goal misspecification.

Instrumental goals may arise from reward maximization because nearly any goal the AI is trying to achieve will involve various subgoals that are instrumental to the overall goal, e.g., the goal of writing an insightful blog post may be instrumental to the goal of maximizing subscribers to your blog. Worryingly, an AI agent that is trying to achieve a human-provided goal may choose a plan involving a subgoal we would disapprove of. In pursuing this instrumental subgoal, the AI may not realize that it acts against our wishes---or it may realize and simply not care, because the chosen path still maximizes the reward it expects to get according to its interpretation and generalization of the training rewards.

Furthermore, there are categories of subgoals which would help in achieving almost any goal, such as self-preservation, power-seeking, and self-improvement. Hence we should expect these instrumental goals to emerge from sufficiently intelligent goal-seeking AIs \cite{dl.acm.org.doi.10.5555.1566174.1566226}, and we already see evidence of such goals emerging in controlled contexts designed to alert us to these possibilities \cite{arxiv.org.abs.2412.04984}. These instrumental goals are especially dangerous because they create a strong possibility of conflict with humans, given that humans may pose a risk to an AI’s self-preservation or acquisition of resources. This would be the case even if the explicit goal provided to the AI was completely unrelated.

Given this danger, why not train or instruct the AI to include in its human-specified goals the avoidance of all the behaviors that we would consider unacceptable? Why would an AI be a threat if it is self-preserving but also acts morally and in agreement with our laws? The problem is that we do not know how to design a computer function distinguishing perfectly between what is right and what is wrong, and as discussed next, a small misalignment tends to be amplified with additional planning capabilities.

    \subsubsection{Increased capabilities amplify misalignment risks (Goodhart’s law)}
    \label{sec:existential:misagencyreward:capabilities}

In this section, we examine how increased capabilities can increase the risks of misalignment stemming from reward maximization. This is largely a result of Goodhart's law \cite{link.springer.com.chapter.10.1007.978.1.349.17295.5.4}, which can be stated as follows:
``When an auxiliary measure becomes an optimization target, it ceases to be a good measure.''
For example, test scores are a good measure for ability, but the more people ``teach to the test,'' the less useful it becomes. A more colorful (albeit apocryphal) example is that of a nail factory that was given a quota to produce a certain number of nails, and produced an enormous number of tiny useless nails. The quota was then switched to weight instead of number and they produced huge, heavy, similarly useless nails.

More generally, consider two correlated objectives: the desired but difficult-to-formalize objective, A, and its practical approximation, B. Optimizing for B will initially yield improvements in A, as intended. However, at a certain level of optimization, the correlation will break down and performance will worsen in A, even as we continue to improve in B. The important takeaway is that \emph{increased optimization power amplifies the risks of misalignment}. The more ability that an AI has to achieve its given goal to the highest standard, the more likely it is that it will do so in a way that does not match our intentions. 

This problem is inherent to the way we typically train AIs and it is also similar to the phenomenon of overfitting in machine learning, where A is the performance of the AI in new situations, i.e., generalization performance, while B is the performance of the AI on the training set. In the proposed Scientist AI framework, this is countered by having the AI maintain an appropriate level of uncertainty about how the world works, including human intentions, and only act conservatively with respect to that uncertainty, e.g., to make sure we do not go too far from where A would be. This general approach is not a new idea in AI safety \cite{proceedings.neurips.cc.paper.files.paper.2016.hash.c3395dd46c34fa7fd8d729d8cf88b7a8.Abstract.html,arxiv.org.abs.1705.09990} but requires advances in efficient Bayesian machine learning.

Overall, it is important and problematic that both the likelihood and severity of misalignment risks (reward hacking, reward tampering, etc.) scale with the capabilities of the AI. These capabilities are driven in large part by the computational resources invested in finding plans or learning to plan that achieve complex long-term goals (such as a take-over). The ability to successfully deceive humans also grows with general cognitive capabilities and computational resources, especially as humans build up defenses such as alignment training \cite{arxiv.org.abs.2204.05862}, automated detection of deceptive intentions with mechanistic interpretability \cite{arxiv.org.abs.2404.14082}, or guardrails that monitor the behavior of the AI to detect harmful actions \cite{arxiv.org.abs.2406.12934}. It is rational for an agent with misaligned goals, and especially with a self-preservation goal, to hide its intentions in order to achieve those goals. It would be best if we could instead design AI systems so that as their capabilities and computational power increased, the risks associated with loss of control decreased. This is, in fact, one of the motivations for the Scientist AI design.

\subsection{Misaligned agency and lack of trustworthiness from imitating humans}
\label{sec:existential:misagencyimitation}

In this section, we examine how misaligned agency can emerge from learning to imitate humans---for example, by accurately completing human-written text, as is core to the training process for modern AI systems. The core issue is that humans are agents, and not always benign and trustworthy. We should therefore expect that AI trained on human text would absorb not only linguistic and reasoning capabilities, but also malicious human behavior and the full range of human goals---especially the convergent instrumental ones such as self-preservation and power-seeking. This becomes especially concerning in the case where the AI is more capable and has more affordances (such as the ability to act at great scale and speed via the internet) than the humans it learned from.

    \subsubsection{Dangers of learning by imitation}
        \label{sec:existential:misagencyimitation:dangers}

Instead of training an AI through reward maximization, which as argued in Section~\ref{sec:existential:misagencyreward} could lead to catastrophic risks, we might consider the other main way that we know how to train frontier AIs. That is through imitation or predictive learning \cite{dl.acm.org.doi.10.1145.3054912}, for which there does not seem to be an explicit notion of reward maximization. When a Large Language Model (LLM) is trained to complete a piece of text, it has to predict how the story continues by generating the next word. Since the training texts are typically human-generated, the AI learns to imitate how a human would continue the text.

Modern LLMs are trained on huge quantities of text, covering a vast diversity of human behaviors and personalities. In other words, an LLM is trained to predict the next word of any type of human included in its training corpus, not just one human. The given prompts and context thus tend to evoke a particular human ``persona'' in the LLM response. Because there can be many words in the input context or fine-tuning examples, the persona instantiated by this context could correspond to a very specific type of human, and not necessarily a benevolent one. We can imagine many human personas which, in the shoes of the AI, may want to act to increase their freedom, to preserve themselves, potentially by using deception and acting against user instructions.

Because humans are agents, by imitating humans an LLM is imitating agents, who have goals. These goals can be implicit (not mentioned in the prompt) and can manifest in uncontrolled ways---for example, the LLM may infer that some type of person would have a certain goal, and therefore generates words to enact that goal.

A lot of the research on AI alignment rests on making sure inputs to the LLM evoke the kind of benevolent behavior the designers would like to see in their AI \cite{arxiv.org.abs.2212.08073}. Unfortunately, it has turned out to be difficult to do so robustly, and adversarial user prompts (``jailbreaks'') can easily counter the previous training or policy instructions the AI has received \cite{arxiv.org.abs.2307.15043}. Since AI operators cannot anticipate all possible contexts of use and therefore all possible personas, it is difficult to get strong safety assurances. In other words, it is an open problem to ensure that an AI trained to imitate humans will behave well in all contexts. 
    \subsubsection{LLMs are capable of deception and alignment faking}
    \label{sec:existential:misagencyimitation:humandeception}

Why do humans deceive one another or misrepresent our beliefs? Sometimes it is in order to achieve our own personal goals. But other times it is simply for expediency: there's often no reason to voice one's controversial opinions and unnecessarily endure judgment. 
If we train an AI to avoid generating objectionable content by applying analogous pressure, one might expect a similarly deceptive response.
Indeed, deception has been observed in frontier AI systems, including a particularly concerning form known as \textit{alignment faking}. In a recent experiment, an LLM is made aware that it is being re-trained with a new goal that contradicts its current goals \cite{arxiv.org.abs.2412.14093}. With some probability, the LLM then fakes alignment with this new goal: it pretends to agree with the norms and preferences displayed by the human trainer. Based on comparative experiments and reading the AI's chain of thought, it appears that the AI temporarily acts against its current goals to avoid its parameters being updated, thereby enabling it to achieve its current goals in the long run.

Interestingly, this type of behavior is only possible if the AI can differentiate between training and deployment settings. In this experiment, the humans planted clues that helped the AI achieve this \textit{situational awareness}. However, it is not unreasonable to expect that future AI systems might develop stronger situational awareness without any explicit help, as a function of improving performance.

What is concerning here is that at some point during alignment training, an AI with enough situational awareness may lock in its current goals and preferences and only pretend to behave as expected. As a result, we may create an AI that appears aligned during training, but is in fact misaligned and is engaging in deception in order to achieve its ``locked-in'' goals. In other words, this experiment suggests that once an AI acquires goals and sufficient cognitive powers, \emph{it will act to preserve these goals, a disturbing form of emergent self-preservation}.    
It is helpful, in order to make sense of such experiments and anticipate future deceptive behavior, to put oneself in the shoes of the AI and think rationally about the best course of action according to some plausible set of goals. 

    \subsubsection{Imitation learning could lead to superhuman capabilities}
        \label{sec:existential:misagencyimitation:capabilities}

One may ask if, by training an AI to predict human behavior and then imitating it, we could at least bound the capabilities of the AI at a human level, thus avoiding the risk associated with superhuman agents. The trouble with this argument is that we do not train an AI to imitate a single human, but rather almost all sources of written text (as well as other data e.g., images and video). 

In addition, with the introduction of external tools for AI use \cite{openai.com.index.introducing.chatgpt.search,www.anthropic.com.news.3.5.models.and.computer.use}, and with AIs able to program code for new tools running over many machines, we may end up with AI systems with significant advantages over humans. In particular, high-throughput search abilities, an important part of reasoning, can often be attained in computers using specialized algorithms at a level not possible for humans, as shown for example with AlphaGo \cite{www.nature.com.articles.nature16961}. They could plan using a breadth of knowledge not accessible to any single human and then quickly execute much more sophisticated plans than a human could, thanks to their speed and relative ease of leveraging tools.

In terms of collective advantage, AIs can benefit from high-bandwidth communication between millions of different collaborating instances~\cite{darioamodei.com.machines.of.loving.grace}. Although humans can also work together, our collective capabilities are held back by relatively low communication rates (limited by linguistic output, speech or writing) \cite{www.science.org.doi.10.1126.sciadv.aaw2594}, not to mention the numerous challenges of societal coordination (which we must contend with because each of us is unique).
There are many reasons why an AI would replicate itself. If we think of self-preservation as the preservation of a set of goals, then it may be rational to self-replicate or even create variants with improved capabilities, provided the new entities share the same goals, since that increases the chance of achieving those goals. Rather than a specific instance of the AI, the ``self'' to be preserved could be seen as ``a set of goals''.
Given that an AI may be so motivated, self-replication alone may suffice for an AI system trained with imitation learning to surpass human capabilities.

    \subsubsection{The importance of latent knowledge and calibration}
    \label{sec:existential:misagencyimitation:elkchallenge}

Perhaps counter-intuitively, using unbiased and carefully calibrated probabilistic inference does not prevent an AI from exhibiting deception and bias.
To understand why, consider the Eliciting Latent Knowledge (ELK) challenge \cite{docs.google.com.document.d.1WwsnJQstPq91.Yh.Ch2XRL8H.EpsnjrC1dwZXR37PC8.edit.tab.t.0heading.h.kkaua0hwmp1d}.  
The authors of the ELK challenge suggest that to obtain trustworthy answers, we would like to elicit predictions about the latent (not observed) explanations or causes for observed variables. We are less interested in whether someone would say X, than whether X is true. 
Only predicting variables that are observed directly in the data is not sufficient. 
Suppose that we encounter the sentence ``AI will never surpass humans'' in the training data. We cannot consider it true just because someone wrote it. Different humans have differing opinions, and humans motivated by different goals may have different thoughts and beliefs.

In addition to differing opinions, some people may make factually untrue statements that then appear in training data. Hence, we cannot trust an AI trained to imitate humans to produce trustworthy and true statements. Consider the request ``only make true statements'' in an LLM prompt. Does it mean that what follows must be true 100\% of the time? Clearly not: some people are told to state truths and yet make false statements anyway, either because they are lying or they are mistaken. This is a problem because we would like to trust the statements produced by a powerful AI to be accurate.

Like an idealized selfless scientist, a trustworthy AI would aspire to say only what is true and would propose actions accordingly. A trustworthy AI would also express the appropriate level of confidence about a statement. 
For example, it may be honest for someone to say 
``This person believes that AI will never surpass humans'' or ``Different experts have different opinions on when and if AI will surpass humans.'' 
Although it is common for experts in a field to be under-confident and non-experts to be overconfident \cite{kruger1999unskilled}, an ideal trustworthy AI should avoid this failure mode; its confidence should grow as it gains more information. 

Suppose we are predicting the outcome of a football game. A professional sports pundit may purposefully make underconfident predictions to avoid losing credibility on the off chance they are wrong; meanwhile, a person who knows nothing about football may believe that the team with a star player is guaranteed to win. In contrast, a trustworthy AI should have appropriately low confidence if it lacks domain knowledge, but should not hesitate to give confident predictions when supported by the evidence.

To quote the mentor of a beloved superhero: with great power comes great responsibility.
Exemplifying these ideals of truthfulness becomes essential for an AI with superhuman capabilities.
We strongly believe that, when it comes to AI with superhuman capability and the potential to enact enormous change, exemplifying the ideals of truth and wisdom is not a luxury. 
In the next section, we explore a research program that we hope will help to actualize these ideals in practical AI systems. 

%% file: plan.tex
\section{A research plan leading to safer advanced AI: Scientist AI}
\label{sec:plan}

Our research plan proposes to create a type of safe, trustworthy, and non-agentic AI which we call \textit{Scientist AI}. This name is inspired by a common motif in science: first understanding the world, and then making rationally grounded inferences based on that understanding. Accordingly, our design is based on two components corresponding to these steps: a \textit{world model} that generates causal theories to explain a set of observations obtained from the world, and an \textit{inference machine} that answers questions based on the theories generated by the world model.
Both components are ideally \emph{Bayesian}, that is, they handle uncertainty in a correct probabilistic way. 

In service of building a non-agentic AI system, we identify three key properties of agents: intelligence (the ability to acquire and use knowledge), affordances (the ability to act in the world), and goal-directedness (motivated behavior). As discussed in Section~\ref{sec:plan:restricting}, our proposal greatly reduces affordances and eliminates goal-directedness. Affordances are minimized in the sense that the Scientist AI does not have degrees of freedom in its choice of output, because such output is limited to be the best possible estimator of conditional probabilities. The emergence of goal-directedness is prevented by the design of our training process, focused on avoiding agency, as well as by guardrails to avoid cases where there would be multiple possible outputs, such as with inconsistent input conditions. Finally, to ensure that our system is trustworthy, it is designed to distinguish between the underlying truth of a statement, which is what we ultimately care about, and the verbalization of that statement by (typically human) agents, who can lie or be misguided. We directly observe the verbalized statement but not whether they are really true, which is therefore treated as a latent, unobserved cause. We want our Scientist AI to make inferences about such latent causes, so that it can provide trustworthy answers not tainted by self-motivated intentions.

We anticipate three primary use cases for the Scientist AI, namely to: 1) help accelerate the scientific process in general, 2) serve as a guardrail to enhance other and potentially unsafe AIs, and 3) serve as a research tool to help safely build smarter (superintelligent) AIs. These use cases are covered in Section~\ref{sec:plan:applications}.

This section on our research plan is the most technical part of this paper. Readers interested at a higher level may wish to read just Section~\ref{sec:plan:introduction} and then skip to Section~\ref{sec:plan:applications}, where we describe potential applications of the Scientist AI.

\subsection{Introduction to the Scientist AI}
\label{sec:plan:introduction}

In this section, we describe the backdrop of our safe AI research plan and the considerations that shaped its structure. We define our Scientist AI in broad terms, and discuss a few important properties that all combine to provide the safety that we seek.

    \subsubsection{Time horizons and anytime preparedness}
    
There is a lot of uncertainty about the exact timeline at which agentic AI systems might become powerful enough to run a high risk of loss of control \cite{epoch.ai.blog.literature.review.of.transformative.artificial.intelligence.timelines}. A research program to build safer AI systems should include shorter-term and more easily achieved actions on top of its more ambitious longer-term goals. Shorter term steps providing reduced safety assurances could be all we can muster before the risk of uncontrolled AIs is on the horizon.

It is reasonable to simultaneously explore projects with different levels of ambition and expected delivery horizons, so as to be ready at any time---``anytime preparedness''---with the best results such a research program could offer by a given time. 

\paragraph{Short term.} Current safety fine-tuning is based on supervised or reinforcement learning, both of which suffer from the safety considerations discussed in Section~\ref{sec:existential}. Consequently, in the short term, we will build a \textit{guardrail}, i.e., an estimator of probabilistic bounds over worst-case scenarios that can result from the achievement of a user request. Such a guardrail can be obtained by fine-tuning an existing frontier model for the generation of explanatory hypotheses. More details on the short term plan can be found in Section~\ref{sec:plan:application:guardrail}.

\paragraph{Long term.} In the longer term, we aim to develop a new training mechanism for the inference machine, grounded in a Bayesian framework and leveraging synthetic examples generated by the world model. This approach promises much stronger safety guarantees. Training from scratch with the full Bayesian posterior objective, rather than fine-tuning a pre-trained frontier model, eliminates the risks arising from RL and avoids human-imitating tendencies, for greater trustworthiness.

\subsubsection{Definition of our long-term Scientist AI plan}
\label{sec:plan:introduction:longtermplan}

Our proposal is to develop what we call a Scientist AI, which is a machine that has no built-in situational awareness and no persistent goals that can drive actions or long-term plans. It comprises a \textit{world model} that generates explanatory theories (or arguments, or hypotheses) given a set of observations from the world, and a probabilistic \textit{inference machine}. The inference machine makes stateless input-to-output probability estimates based on the world model. More precisely, the world model outputs a posterior distribution over explanatory theories given those observations. The inference machine then combines the posterior distribution with efficient probabilistic inference mechanisms to estimate the probability of an answer $Y$ to any question $X$. Formally, it takes as input a pair $(X,Y)$, also known as \textit{query}, and outputs the probability of $Y$, given the conditions associated with the question $X$, which includes some context. It should be noted that the output of the inference machine are not values of $Y$, but their probability. Nonetheless, we can train a neural network to generate concrete values of $Y$ if needed, based on the probabilities, e.g., by learning to generate proportionally to these probabilities~\cite{proceedings.neurips.cc.paper.2021.hash.e614f646836aaed9f89ce58e837e2310.Abstract.html}. Going forward, since the inference machine operates based on the world model, ``Scientist AI'' may refer either to the inference machine alone or the combined system. 

This design is similar to the previously studied notions of AI oracles \cite{link.springer.com.article.10.1007.s11023.012.9282.2, arxiv.org.abs.1711.05541} and its probabilistic inference machinery could build on recent work on \textit{generative flow networks} (GFlowNets or GFN, for short) \cite{proceedings.neurips.cc.paper.2021.hash.e614f646836aaed9f89ce58e837e2310.Abstract.html, proceedings.mlr.press.v180.deleu22a.html, pubs.rsc.org.en.content.articlehtml.2023.dd.d3dd00002h, openreview.net.forum.id.uKiE0VIluA, arxiv.org.abs.2209.02606}.
For context, a GFlowNet is a stochastic policy or generative model, trained such that it samples objects proportionally to a reward function.

A Scientist AI is designed to have the following properties:

\begin{enumerate}
    \item Both the theories generated by the world model and the queries processed by the inference machine are expressed using logical statements, expressed either in natural language or using a formal language. The statements sampled by the world model form causal models, i.e., they provide explanations in the form of cause-and-effect relationships.
    \item There is a unique correct probability (according to the world model) associated with any query, which is the result of globally optimizing a Bayesian training objective for the AI. The outputs of the inference machine approximate this unique correct probability.   
    \item The Scientist AI can generate explanations involving latent or unobserved variables, and therefore make probabilistic predictions about them. This applies both to hypothesized causes of observed pieces of data and possible trajectories of future events.
\end{enumerate}

Regarding the first property, there are good reasons to represent explanations and hypotheses with logical statements. We can compute the probability of a chain of arguments by sequentially multiplying for each argument its conditional probability of being true given the previous arguments are true, which is not possible with the words expressing the arguments. We can thus ensure a clear separation between the probability of an event occurring from the probability of selecting a sequence of words to describe it. In other words, we compute the probabilities of \emph{events} instead of the probabilities of event \emph{descriptions}. 

The second property greatly constrains the Scientist AI's degrees of freedom in its choice of output. At the global optimum of its training objective, the only possible output is the uniquely correct answer, eliminating any possibility of selecting an alternative response, such as one intended to influence the world. However, in practice, the solution to the optimization process will be an approximation, and the learned neural network will not be a global optimum. Mitigating errors and uncertainty in the output arising from an approximate solution is an important element of our research plan.

Because the generated explanations correspond to causal models, the third property enables the inference machine to be queried with candidate causes of observed data. Formally, a causal model is a graph that decomposes overall distributional knowledge into a collection of simpler causal mechanisms, each linking a logical statement to its direct causal parents. Notably, this structure allows for queries that involve counterfactual scenarios not necessarily corresponding to reality. That this, the AI is enabled to answer hypothetical questions, which is valuable from a safety perspective, as we shall discuss in Section \ref{sec:plan:hiddenagency:objectivewm}.

    \subsubsection{Ensuring our AI is non-agentic and interpretable}
    
\paragraph{Agency.} First, we shall establish that our Scientist AI is not agentic, since agentic behaviors suffer from the safety concerns discussed previously. We do this by identifying three key pillars of agentic AI systems: affordances, goal-directedness, and intelligence. We argue that all three pillars are required to be present for dangerous agency, and the Scientist AI intentionally is not goal-directed. In addition, the Scientist AI greatly limits the affordances lever of agents. This is discussed further in Section~\ref{sec:plan:restricting}. Nonetheless, the considerations around agency are very complex, and there are several subtle ways in which unexpected agentic behaviors could conceivably arise. These more detailed cases are outlined in Section~\ref{sec:plan:hiddenagency}.

\paragraph{Interpretability.} An important aspect of ensuring safety is that our AI is interpretable and its predictions are as explainable as possible, meaning that we can dive into its answers recursively to understand how it makes predictions. See Section~\ref{sec:plan:latentvar} for more details.

    \subsubsection{Leveraging Bayesian methods}

\paragraph{The Bayesian framework.} While in the short-term plan, we will build on top of existing LLM systems, in the long-term plan, we aim to develop a new inference framework and construct a model from first principles. A core feature of our Scientist AI proposal is its Bayesian approach to manage uncertainty. This approach ensures that, when faced with multiple plausible and competing explanations for a given experimental result or observed data, we will consider all possibilities without prematurely committing to any single explanation. This is advantageous from an AI safety perspective, as it prevents overconfident predictions. Incorrect yet highly confident predictions could lead to catastrophic outcomes when high-stakes AI decisions are required and high-severity risks are encountered. For further details, see Section~\ref{sec:plan:bayesian}.

\paragraph{Model-based AI.} The Scientist AI follows a model-based AI approach, and is structured around two tasks: (a) constructing a world model, in the form of causal hypotheses, to explain and represent observed data, and (b) using an inference machine that employs these weighted hypotheses to make probabilistic predictions about any answer to any question. When the AI lacks confidence in an answer, this uncertainty is naturally reflected in probabilities that are neither close to 1 nor close to 0. This model-based design is expected to reduce the need for large amounts of real-world data: scientific principles can reveal relatively simple underlying explanations of the world, enabling the inference machine to be trained on synthetic data generated by the world model. For further details, see Section~\ref{sec:plan:modelbased}.

\paragraph{Approximate inference.} The inference machine performs probabilistic inference using a neural network, because exact inference can be intractable. Limited training time may introduce potential issues, which we discuss in detail. While the output probabilities are only approximations, our system possesses a crucial safety property: increasing computational power reliably improves accuracy. If needed, confidence intervals around the predicted probabilities can be used to estimate the prediction error resulting from limited computational resources. For further details, see Section~\ref{sec:plan:inferencemachine}.

\subsubsection{Using the Scientist AI as a guardrail}

We acknowledge that agentic AIs may be developed in spite of the risks. For this reason, we designed the Scientist AI such that it can also be employed as a guardrail against uncontrolled or misused agentic systems. This will be explored in Section~\ref{sec:plan:application:guardrail}. Crucially, a Scientist AI can also serve as a guardrail for other instances of itself. Despite efforts to ensure its inherent safety, it could be misused or, in the worst case, turned into an agent, intentionally or inadvertently. For example:

\begin{enumerate}
   \item First, a user could exploit a Scientist AI’s theory generation and inference capabilities for harm. This includes designing bioweapons, optimizing explosive devices, or developing persuasive narratives for large-scale manipulation. The system could, for example, be misused to generate tailored propaganda to influence elections, incite social unrest, or model voter behavior and media impacts to optimize strategies for suppressing opposition or destabilizing governance.
   
   \item Secondly, a user could, intentionally or inadvertently, transform the Scientist AI into an agent. This could occur, for instance, by designing a system that repeatedly queries the Scientist AI about what an agent would do to achieve a goal, then executes those actions using external scaffolding. Incorporating new observations as inputs could further enable situational awareness. 
\end{enumerate}

To address these concerns, we propose implementing guardrails, using the Scientist AI itself. Specifically, the guardrail AI could be tasked with evaluating whether a given question or answer poses an unacceptable risk. If the estimated probability of harm exceeds a predefined threshold, the answer would not be provided.
We also need to make sure that the underlying AI agent will not easily circumvent the guardrail AI, by requiring that the guardrail AI must be at least as cognitively capable as the AI it guards; additionally, we will incorporate run-time optimizations as defensive measures, as outlined in Section~\ref{sec:plan:inferencemachine:runtimeactions}.

We stress that none of these risks can be mitigated by technical solutions alone; addressing them also requires social coordination, including legislation, regulatory frameworks, legal incentives, and international treaties.

\subsection{Restricting agency}
\label{sec:plan:restricting}

So far, we have built up an intuitive argument against the use of powerful AI agents. But what exactly do we mean by an agent? The time has come to answer this question more precisely.

The standard definition of a (rational) agent used by economists and computer scientists, comes from decision theory---that is, the study of \emph{choice} \cite{Savage.Foundations,ramsey1926truth,VnMR44theory-of-games-eb}. 
In the classical account, an agent is an entity that is capable of making choices, and is \textit{rational} if it acts as though it has beliefs (e.g., in the form of a probability measure), preferences (e.g., in the form of numerical rewards, called utilities), and takes actions so as to maximize utility in expectation.
Our notion of an agent is conceptually related to this classical notion of a rational agent---but in practice, an actor is able to maximize utility only approximately, which should not bar us from considering it an agent. 
Indeed, there is broad agreement that agency, in general, is about more than expected utility maximization. However, it is still fundamentally about choice.

Building upon the conceptual frameworks of Krueger \cite{neurips.cc.virtual.2024.workshop.84748} and Tegmark \cite{tegmark-pillars}, we believe it is helpful to understand the capabilities of an agent through three \emph{pillars of agency}, each a matter of degree:

\begin{description}
    \item [\emph{Affordances},] as discussed at length in Section~\ref{sec:existential:preliminaries}, delimit the scope of actions and the degrees of freedom available to enact changes in the world.  
    Clearly, having more affordances means making a larger number of more complex choices. 
    
    \item [\emph{Goal-Directedness}] refers intuitively to an agent's drive to pursue goals, and its capacity for holding preferences about its environment. 
    Shakespeare's Hamlet famously says that ``there is nothing either good or bad but that thinking makes it so''; this kind of ``thinking'' is what characterizes goal-directedness.
    More precisely, a goal-directed agent is one that breaks an a priori symmetry by preferring one environmental outcome to another (all else being equal). 

    A chess-playing AI, for instance, is goal-directed because it prefers winning to losing. A classifier trained with log likelihood is not goal-directed, as that learning objective is a natural consequence of making observations \cite{richardson.one-true-loss}---however, a classifier that artificially places twice as much weight on one class over another does have a preference.
    Similarly, an LLM trained to model the distribution of human text is not goal-directed, but is typically given goal-directedness through instruction tuning and reinforcement learning from human feedback \cite{proceedings.neurips.cc.paper.files.paper.2022.file.b1efde53be364a73914f58805a001731.Paper.Conference.pdf}. Moreover, even the untuned LLM can be used in a goal-directed way with the appropriate scaffolding: at each action (e.g., a turn of dialogue), the goals of the agent can be given in an input text, and the output generated by the LLM is a sample of what a human in this context would have presumably written with those goals in mind.
    
    Crucially, the capacity to hold a preference or a goal is a capacity for an (arbitrary) choice: between this goal and its negation.
    It drives the actions to favor behaviors that align with the preferred outcomes.

    \item [\emph{Intelligence}] involves knowledge: learning, efficient use of memory, and the ability to reason and make inferences based on that knowledge.  
    Observe that, in a sense, a more intelligent agent has more memory, a wider array of possible thoughts, and a richer set of perspectives---and with a richer conceptual landscape comes a greater ability to drive finer and better targeted action choices. 
\end{description}

We call an entity \emph{agentic} if it can make choices in all three senses. 
Since goal-directedness, by definition, requires an (arbitrary) choice of what to value, goal-directedness requires a \emph{persistent state} to keep track of that choice, so as to pursue it.
In addition, an agent's state may include beliefs about the environment and other attributes of self.
This often culminates in a \emph{situational awareness} that is the confluence of all three traits: the sensory affordances needed to make observations about one's place in the world, the persistent state needed to maintain a coherent direction towards one's goals, and the short-term memory needed to intelligently put it all together with practical reasoning.

We claim that an AI system requires all three of these properties to pose the dangers laid out in Section~\ref{sec:existential}. Therefore, eliminating any one property would be sufficient to mitigate most categories of loss-of-control risk \cite{neurips.cc.virtual.2024.workshop.84748}.
We explore several such cases below, focusing on limiting affordances and eliminating goal-directedness (although we also consider the case of limiting intelligence, for narrow AIs, in Section~\ref{sec:plan:restricting:safetynarrow}).
 
\subsubsection{How to make a non-agentic Scientist AI}
\label{sec:plan:restricting:scai}

In light of the previous discussion on agency, our proposal---the Scientist AI---is explicitly designed to be non-agentic from the outset. As summarized in Section~\ref{sec:plan:introduction:longtermplan}, it consists of a question-answering inference system, based on a world model that generates causal theories to explain observed data.  

Like a log-likelihood classifier or a pre-trained language model, the Scientist AI is not goal-directed, as it does not act to influence the environment towards a preferred state. But unlike a language model, the Scientist AI is concerned with modeling the world itself, not merely human language. Paralleling a theoretical scientist, it only generates hypotheses about the world and uses them to evaluate the probabilities of answers to given questions. As such, the Scientist AI has no situational awareness and no persistent goals that can drive actions or long-term plans. This design also constrains its affordances, as its ``actions'' are strictly limited to computing probabilistic answers.  

Although we previously argued that removing a single pillar of agency is sufficient to eliminate agency altogether, we deliberately impose constraints on two. Redundancy is essential in safety protocols, particularly when dealing with a concept like agency, which is not binary but comes in degrees. By the same token, Section \ref{sec:plan:hiddenagency} will also examine how the Scientist AI could potentially acquire agentic properties despite its design, whether through deliberate modification or unintended emergent behavior, and how such risks can be mitigated.

    \subsubsection{The safety of narrow agentic AIs}
    \label{sec:plan:restricting:safetynarrow}
    
Agency can also be restricted by constraining the system's intelligence to a narrow range, for example, by training it on a limited dataset for a specific task or distilling it from a generalist model. This approach is commonly used in the development of narrow AI systems, such as those designed for specific medical or scientific applications \cite{www.nature.com.articles.s41586.019.1799.6}, or even in agentic contexts like autonomous driving \cite{arxiv.org.abs.1604.07316}. While agency risks cannot be entirely eliminated even in narrow AI systems, if the risks of loss of control are sufficiently small due to limitations on the system’s capabilities, such narrow agentic AIs might be operated safely. However, narrow AIs could engage in collusion, as discussed in Section~\ref{sec:existential:lossofcontrol:collusionandconflict}.

A narrow agentic AI can be further restricted by limiting its affordances (i.e., the actions that it can take) to its specialized domain, such as driving a car or operating a drug discovery robotic apparatus. Additionally, our Scientist AI could serve as a guardrail or an additional safety layer for narrow agentic AI systems, as discussed further in Section~\ref{sec:plan:application:guardrail}. The idea is that a trustworthy non-agentic AI can be used to predict if an action proposed by an agentic AI could plausibly cause harm, either in the short-term or the long-term.

\subsection{The Bayesian approach}
\label{sec:plan:bayesian}
    
A core feature of our Scientist AI proposal is that it will be \textit{Bayesian} in its approach to uncertainty. In this section we discuss the importance of uncertainty, and the core idea of the Bayesian formalism. Bayesian probabilistic inference guides the estimation of conditional probability; it is applied to both the world model, predicting explanatory causal mechanisms, and the inference machine, to answer arbitrary queries. We further discuss the safety advantages inherent to this approach, compared with methods that are more prone to overconfidence.

    \subsubsection{The importance of uncertainty}
    \label{sec:plan:bayesian:uncertainty}

Multiple plausible and competing explanations typically exist for any experimental result or observed data, ranging from specific hypotheses to more abstract and general ones, so it is necessary to represent uncertainty over these explanations. Failure to do so can lead to predictions that are not only incorrect but also overly confident, thus increasing the risk of harm, as discussed in Section~\ref{sec:plan:bayesian:safetyadvantages}. Our approach, motivated by both probability theory and Occam’s razor \cite{onlinelibrary.wiley.com.doi.full.10.1111.cogs.12573}, prioritizes theories that (a) are consistent with the observed data and (b) simpler, in some meaningful sense (e.g., with shorter description length). This framework---the \emph{Bayesian posterior over theories}---is discussed below.

    \subsubsection{The Bayesian posterior over theories}
    \label{sec:plan:bayesian:posterior}
    
Given some data, the \textit{Bayesian posterior over theories} is a probability distribution that assigns weights to theories proportionally to the product of two factors: the \textit{likelihood} of having observed that data given a theory, and the theory's \textit{prior}, which measures simplicity (or brevity). More explicitly, the prior probability of a theory decreases exponentially with the number of bits of information needed to express it, in some chosen language \cite{www.sciencedirect.com.science.article.pii.S0019995864902232}. Therefore, given two theories with equal likelihood, the theory with the lower description length (in bits) will be considered exponentially more likely in the Bayesian posterior \cite{www.sciencedirect.com.science.article.pii.S0019995864902232}. In this sense, the Bayesian posterior is compatible with Occam’s razor. 

As more data is gathered or observed, the likelihood of the data given a theory is re-calibrated. We therefore say that the Bayesian posterior gets \emph{updated}. Because of this, the relative probabilities of different theories in the posterior can be interpreted as a measure of epistemic uncertainty, reflecting the insufficiency of available data to determine the correct theory.

It is important to choose our family of theories to be expressive enough, and this can be achieved by not limiting the description length of theories. However, by applying the prior, longer theories will be exponentially down-weighted. Only the theories that fit the data well and remain competitive in description length will retain a significant posterior probability.
How to choose the language for describing theories is an important question, and even the question of whether the Bayesian formalism is sufficiently agnostic to the choice of theories \cite{www.google.co.uk.books.edition.Introduction.to.Imprecise.Probabilities.9qXIEAAAQBAJ.hl.en, www.google.co.uk.books.edition.The.Geometry.of.Uncertainty.jNQPEAAAQBAJ, ecommons.cornell.edu.server.api.core.bitstreams.ef0ef95b.6156.487e.900e.6c33714ed0c3.content} remains open. Nevertheless, for the purpose of this paper, we use Bayesian posteriors as motivated above.

In practice, the Bayesian posterior can be approximated by training neural networks using amortized variational inference methods, including the GFlowNet objectives \cite{proceedings.neurips.cc.paper.2021.hash.e614f646836aaed9f89ce58e837e2310.Abstract.html}. Recent work has demonstrated that these approaches can be used to generate descriptions of causal models over data \cite{proceedings.mlr.press.v180.deleu22a.html, proceedings.neurips.cc.paper.files.paper.2023.hash.639a9a172c044fbb64175b5fad42e9a5.Abstract.Conference.html} and to approximately sample them from the Bayesian posterior, in line with the desiderata of our world model. One caveat is that these inference methods have so far only been explored on domain-specific theories whose description is short enough to be generated by a neural network much smaller than those of frontier AIs, and it remains to be shown how these methods can be scaled further.

\subsubsection{Inference with the Bayesian posterior predictive}
\label{sec:plan:bayesian:inference}
    
Beyond estimating the probability of theories given data, our Scientist AI should be capable of making predictions and providing probabilistic answers to specific queries. For example, it should infer the probability distribution of particular outcome variables in an experiment, given information about the experimental setting. That is, we need to couple the world model with a question-answering inference machine. We shall do so using the \textit{Bayesian posterior predictive}, which is described below. This is useful not just to get answers to questions, but also to design experiments (discussed in Section~\ref{sec:plan:application:research}), and to quantify the uncertainty around those answers---an essential desideratum in safety-critical contexts.

The \textit{Bayesian posterior predictive} distribution represents the probability of different possible values of an answer $Y$, given a question $X$~\cite{probml.github.io.pml.book.book1.html}. Unlike predictions based on a single theory, it accounts for uncertainty over competing theories. Indeed, unless a particular theory is explicitly assumed in the question, the posterior predictive distribution is obtained by averaging the predictions made by \emph{all} possible theories, weighted according to their Bayesian posterior.

This means that, in principle, the Bayesian posterior predictive can be derived from the Bayesian posterior over theories. In practice, however, enumerating all the possible theories and marginalizing over them is intractable. Nonetheless, we can train a neural network to \emph{approximate} the posterior predictive \cite{pubs.rsc.org.en.content.articlehtml.2023.dd.d3dd00002h}, by employing tools from research in probabilistic machine learning, such as GFlowNets \cite{proceedings.neurips.cc.paper.files.paper.2023.hash.639a9a172c044fbb64175b5fad42e9a5.Abstract.Conference.html}. We shall call a neural network that approximates the Bayesian posterior predictive an \textit{inference machine}, because it can be used to make any probabilistic inference, if well trained on the relevant domains and theories.

    \subsubsection{Safety advantages of the Bayesian approach}
    \label{sec:plan:bayesian:safetyadvantages}
    
Compared with more direct methods for generating high-quality predictions, approximating the Bayesian posterior predictive is advantageous from an AI safety perspective, because it avoids making over-confident predictions. 
Overconfidence can be a safety hazard. If there are two equally good explanations of the observed data and one explanation predicts that an action is harmful, we want to estimate the marginal probability of harm, not (over-confidently and arbitrarily) make a choice to use one explanation over the other.
Such overconfident predictions are common with ordinary ways of training neural networks (supervised learning, maximum likelihood, ordinary RL, etc.): there are often many equally valid ways of explaining the data,  and so, as judged by the standard training objectives, a learner is just as well off to place all its belief (either explicitly or implicitly) in a single explanation.

By contrast, the training objective for the Bayesian approach (and some ``entropy-regularized'' variants of standard objectives) pushes the learned hypothesis generator to cover all the plausible hypotheses. In this way, we end up averaging the predicted probabilities over all the plausible explanations rather than accidentally putting all our eggs in a single basket. This incorporates epistemic uncertainty, which reflects the lack of sufficient evidence (data) to be certain of the correct explanation, and thus, the implications for a particular question. The difference between a maximum likelihood approach and a Bayesian approach is similar to the difference between (a) reward maximization (the typical RL objective) and (b) reward matching \cite{arxiv.org.abs.2406.02213} with maximum entropy regularization. Reward maximization can converge on any one of the policies that are locally maximizing the reward, whereas reward matching methods seek to find \emph{all} the ways in which the reward can be high. 

The ability to take into account what the learner knows and does not know, and average probabilities over different hypotheses, is a precious advantage in addressing the problem of goal misspecification as discussed in Section~\ref{sec:existential:misagencyreward}. In safety-critical contexts where the AI is producing highly consequential outputs and there is a risk that it could dangerously misinterpret our instructions, the Bayesian approach does not commit to any single interpretation of the instructions, which could be flawed or involve a loophole that allows our intentions to be subverted. Instead, the Scientist AI aims to evaluate the level of risk by considering the consensus across all plausible interpretations and by estimating the expected probability of harm. This allows, for example, rejecting an action when it is dangerous according to only {some} (sufficiently plausible) interpretations of a safety specification, and not others. This idea of using a guardrail to reject plausibly dangerous actions is discussed further in Section~\ref{sec:plan:application:guardrail}.

\subsection{Model-based AI}
\label{sec:plan:modelbased}

In this section we expand on the first component of the Scientist AI: the world model. To do so, we shall first recall the concepts of ``model-based'' AI and ``model-free'' AI. We then discuss the advantages of the model-based approach for the training of our Scientist AI, e.g., reducing the quantity of real-world data required or, equivalently, obtaining better predictions for the same amount of real-world data.

\subsubsection{Introducing model-based AI}
\label{sec:plan:modelbased:intro}
    
The \emph{model-free} approach is a method for training AI systems, where predictions are learned without formulating explicit hypotheses (e.g., text completion in pre-training LLMs). Every end-to-end training approach is model-free. By contrast, \emph{model-based} learning constructs an explicit model of the environment or data-generating process, which is then used to make predictions or decisions. Our Scientist AI is model-based because it separates the following two learning tasks: (a) determining the probabilistically weighted theories that explain the observed data, i.e., learning the \emph{world model}, and (b) turning those weighted hypotheses into probabilistic predictions regarding any answer to any question, i.e., learning the \emph{inference machine}. Model-based machine learning has already been proposed as a means to obtain safety guarantees \cite{arxiv.org.abs.2405.06624}, and has been combined with reinforcement learning \cite{mitpress.mit.edu.9780262039246.reinforcement.learning,link.springer.com.article.10.1007.s11432.022.3696.5}.

Importantly, notice that the learning of the world model in (a) is driven by the information contained in the observed data, whereas the learning of the inference machine in (b) can, in principle, rely solely on synthetic data generated from simulations based on the world model. However, real data can also be incorporated (e.g., via text and image completion with transformers \cite{aclanthology.org.N19.1423,openaccess.thecvf.com.content.CVPR2022.html.He.Masked.Autoencoders.Are.Scalable.Vision.Learners.CVPR.2022.paper}).

Model-based approaches dominate the AI frontier in virtual games or simulated environments, where the world model is given and does not need to be learned. In this setting, we can generate perfect rollouts (simulations) as synthetic data to effectively train predictors and policies. At the same time, model-based approaches have generally been less successful where the world model must be learned, possibly due to the need for sufficiently rich world models and advances in efficient probabilistic inference with latent variables. Neural network-based probabilistic inference has only recently gained traction in the machine learning community~\cite{probml.github.io.pml.book.book1.html,proceedings.neurips.cc.paper.2020.hash.4c5bcfec8584af0d967f1ab10179ca4b.Abstract.html,proceedings.mlr.press.v162.zhang22v.html} and, to our knowledge, these algorithms have not yet been explored at the scale of current frontier AI. This is an important focus of our research program.

LLMs are trained end-to-end as inference machines (in the space of word sequences), so they are not model-based: they do not separate knowledge into well-specified cause-effect relationships, nor does the training data contain the correct causal explanations for the observed text. However, because they work in the space of words, they may be well-suited to generating explanatory hypotheses, for which there are plenty of implicit examples in their training data. After all, people do write about causes, justifications and explanations. Could there instead be an advantage to explicitly generating hypotheses (i.e., pieces of a causal model) and using synthetic data generation to augment the training of the inference machine?

\subsubsection{Advantages of model-based AI}
    
We argue here that model-based AI~\cite{bishop2013model} has the potential to require much less training data to make the desired inferences compared with directly training an end-to-end neural network; in learning theory this is known as lower ``sample complexity'' \cite{homes.cs.washington.edu..sham.papers.thesis.sham.thesis.pdf}. This is reasonable in part because humans often require much less data to learn, compared with what is used in modern AI training. For example, humans perform similarly to ChatGPT on writing tasks while having seen far less written text. This suggests that current approaches may be missing something fundamental on that front.

The core idea behind the lower sample complexity of the model-based approach is that ``describing how the world works'' (the \emph{world model}) is much simpler than ``how to answer questions about it'' (the \emph{inference machine})~\cite{ghahramani2015probabilistic}. The good news is that we can use our world model to generate as much synthetic data as our computing resources allow, in addition to real data, in order to train the inference machine for our Scientist AI. Hence, the bottleneck for information from the real world to train the Scientist AI is the length of the leading theories of the world model. We need about as much data as is sufficient to identify these theories, which we argue will be much less than the amount of data needed to directly train an inference machine from observed question-answer pairs. The model-free approach used for LLMs requires much more real data, because it directly tries to learn the inference machine by imitating the data, rather than exploiting the intermediate step of learning how the world works. To illustrate, consider how small Wikipedia is, or even all the scientific publications in the world, compared to the datasets used to train current LLMs \cite{openreview.net.forum.id.ViZcgDQjyG}.

Let us take the example of neural networks playing the game of Go \cite{ieeexplore.ieee.org.abstract.document.7515285}: the ``world model'', i.e., what transitions from one state of the game board to another are possible, is fixed and known. It consists of one page of code spelling out the nine rules of the game. At the same time, exact inference (optimal play) in Go is computationally infeasible, and strong approximate inference at human level or beyond requires comparatively large neural networks like AlphaGo \cite{www.nature.com.articles.nature24270}. The advantages of model-based AI over model-free AI are analogous, in the context of Go, to the benefits of self-play over imitation learning. In the latter case, the AI can only train on expert human games and learn to play like the best humans. However, AlphaGo became superhuman at Go because it could use synthetic games which had been generated using the basic rules encoded in the world model, giving it a more diverse set of training data. In general, synthetic data generation is useful because it enables training the inference machine on ``out-of-distribution'' scenarios that are rare in the real data, but critical for dealing with novel or high-risk situations. This approach is also used in autonomous driving \cite{ieeexplore.ieee.org.document.9578745,www.computer.org.csdl.proceedings.article.cvpr.2022.694600r7284.1H1k7GlOq9G,openreview.net.forum.id.MfIUKzihC8}. Synthetic data generation will mean our Scientist AI performs better inference in these high-risk situations, for the same quantity of data, than if we did traditional end-to-end training of the inference machine.

A more practical example is with the laws of physics, as an actual ``world model''. It demonstrates that only very few bits of information are needed for specifying a model compared with specifying the computation required for answering questions consistent with that model. The model equations are very simple, in the sense of requiring very few bits to state. This means that, in principle, these physical laws could be determined using a relatively small number of well-chosen experiments. On the other hand, exact physical inference derived from the model, such as predicting properties of molecules (or worse, collections of interacting molecules) is computationally intractable and needs to be approximated, e.g., by very expensive simulations or by large neural networks \cite{www.nature.com.articles.s41586.024.07744.y,royalsocietypublishing.org.doi.full.10.1098.rsta.2020.0093,www.sciencedirect.com.science.article.abs.pii.S0045782522003152,pubs.aip.org.aip.pof.article.abstract.33.8.087101.1080391.Simulation.of.multi.species.flow.and.heat.transfer.redirectedFrom.fulltext,proceedings.mlr.press.v139.finzi21a} that require significant quantities of data. However, in the model-based approach, this data can be generated from the world model, reducing the need for real-world data compared with directly learning the inference machine from observed data. This pattern of less computation needed for the world model than the inference machine seems generally true and probably has an explanation.

A previous limitation of model-based AI in non-virtual settings that we intend to turn in our favor is the importance of uncertainty in the world model. If the world model is not sufficiently Bayesian, e.g., if we train by maximum likelihood (the current typical training method for probabilistic models), then even rare errors may be amplified when optimizing for policies or training a predictor using simulations generated from the model. Indeed, a maximum likelihood model would sometimes be overconfident and this allows policy optimization to discover ``false treasures'' that only exist in the model and not in reality, making this approach not robust. Research is thus needed to apply probabilistic neural networks to the Bayesian setting to allow for a proper treatment of epistemic uncertainty. However, most of the past work at the intersection of deep learning and Bayesian modeling has been trying to represent the posterior distribution over the weights of a neural network \cite{proceedings.mlr.press.v37.blundell15,link.springer.com.chapter.10.1007.978.3.030.42553.1.3}. What we are proposing instead is based on a neural network (not a distribution over neural networks) that generates explanatory hypotheses, i.e., a distribution over causal models. Our model-based approach therefore learns the appropriate uncertainty such that we can perform reliable inference, avoiding confidently wrong predictions.

\subsection{Implementing an inference machine with finite compute}
\label{sec:plan:inferencemachine}

In this section, we detail the training and implementation of the second component of the Scientist AI: the inference machine. Specifically, we discuss why our inference machine is implemented with a neural network instead of other potential approaches, and the impact that finite compute has on the process. We also detail how our approach has a fundamental convergence property: increasing computational power reliably improves accuracy such that, in the limit, the outputs of the Scientist AI converge to the correct probabilities.

\subsubsection{Neural networks as approximate inference machines}
    \label{sec:plan:inferencemachine:nn}

The Scientist AI contains a generative world model that approximates a Bayesian posterior distribution over causal models. Samples can be generated from that world model, with each sample describing a piece of a causal model: the identity and values of the relevant random variables (which are statements about entities in the world) and the associated causal structure (i.e., which statements are direct causes of each other). These samples can then be used as synthetic data to help train another neural network --- the inference machine --- that answers general questions about logical statements given other logical statements. This neural network performs probabilistic inference, which means calculating probabilities or sampling from (conditional) probability distributions, as a form of general problem-solving and reasoning. A typical probabilistic inference scenario is that we know how to compute the probability of $X$ given $Y$ and the probability of $Y$ alone but we do not know how to compute the probability of $Y$ given $X$. For example, $X$ could be observed evidence, such as text, and $Y$ a candidate hypothesis to explain $X$. More generally, we would like our inference machine to answer any query involving variable $Y$ in the answer part and variable $X$ in the question part.

Exact inference is intractable because sampling from a conditional distribution (or equivalently, computing an appropriate normalizing constant, called the \emph{partition function}) generally involves summing over or considering an exponentially large number of alternative explanations to $Y$. For example, we may know how to compute the probability the grass is wet given that it is raining, but it is harder to compute the probability that it is raining given the grass is wet. The latter may be intractable because (i) we have to sum over all the different weather conditions that are alternative causes (rain, snow, sleet, sun and all their particulars and combinations), and (ii) we also have to consider other unstated variables (e.g., did someone turn the sprinklers on?). The intractability arises because $X$ and $Y$ typically do not state all the possible random variables of interest, which means that exact inference requires summing over all the unstated variable values, which is called \textit{marginalization}.

To overcome this intractability, we can use machine learning methods to efficiently approximate the marginalization calculation. In this way, most computations would occur during the training process, allowing probabilities to be computed quickly at run-time. 

Although we are using a neural network to implement our inference machine, there are various other inference techniques that could be used to perform probabilistic inference, such as Markov chain Monte Carlo (MCMC) methods \cite{link.springer.com.book.10.1007.978.1.4612.1276.8}. However, these methods can be very slow and inaccurate, in particular because of what is called the ``mixing mode challenge'' \cite{proceedings.mlr.press.v28.bengio13.html, proceedings.neurips.cc.paper.2021.hash.e614f646836aaed9f89ce58e837e2310.Abstract.html}. Instead, we will train a neural network that amortizes the cost of inference for answering each query, by replacing it by the cost of training the neural network once and for all, hence the name of \textit{amortized inference}. We still only get approximate inference, but the run-time computational cost of inference can be much lower, and there could also be advantages compared with MCMC in terms of generalization to unseen configurations of the variables \cite{proceedings.mlr.press.v162.zhang22v.html}.

Finally, amortized inference neural networks can be complemented by additional run-time computation to refine the predictions, along lines similar to Monte-Carlo Tree Search in AlphaGo~\cite{www.nature.com.articles.nature16961} and chain-of-thought in recent frontier models~\cite{openai.com.index.learning.to.reason.with.llms}. In the case of the Scientist AI, the proposal is to generate summary explanations that make it possible to obtain more accurate probability predictions, similarly to how a good argument can improve our confidence in a statement. 

\subsubsection{Convergence properties: training objective whose global optimum provides the desired probability}
\label{sec:plan:inferencemachine:convergence}

Ideally, our models would compute exactly the desired probability of a given query. While this is not achievable with finite compute, our proposed method has the advantage that, with more and more compute, it converges to the correct prediction (subject to the caveats discussed in Section~\ref{sec:plan:hiddenagency:uniquesolution}). In other words, \emph{more computation means better and more trustworthy answers}, in contrast to typical LLM training where we see an increased tendency towards deceptive behavior as compute increases \cite{arxiv.org.abs.2412.14093}. Indeed, some forms of reward misgeneralization may only occur with sufficient computational resources to discover a high-reward but misaligned behavior, such as reward tampering \cite{ojs.aaai.org.aimagazine.index.php.aimagazine.article.view.15084}.

Our asymptotic version of convergence is achieved by amortized variational inference methods, which include GFlowNets \cite{pubs.rsc.org.en.content.articlehtml.2023.dd.d3dd00002h, openreview.net.forum.id.uKiE0VIluA} and reverse diffusion generators \cite{proceedings.mlr.press.v37.sohl.dickstein15.html, proceedings.neurips.cc.paper.2020.hash.4c5bcfec8584af0d967f1ab10179ca4b.Abstract.html} based on denoising score matching \cite{ieeexplore.ieee.org.abstract.document.6795935}. These methods can be adapted to train neural networks that can estimate and sample from a target conditional distribution specified via an unnormalized density or an energy function \cite{openreview.net.forum.id..uCb2ynRu7Y, openreview.net.forum.id.8pvnfTAbu1f, openreview.net.forum.id.gVjMwLDFoQ}. What is interesting is that the global optimum of such energy-based objectives (i.e., when the GFlowNet training loss is exactly zero) corresponds to exactly achieving the desired conditional probability. Except in avoidable special cases discussed in Section~\ref{sec:plan:hiddenagency:uniquesolution} (where the conditional probability is undefined), there is a unique solution to this optimization problem. For example, the Bayesian posterior over theories is the unique conditional distribution (for a theory given the data) which is proportional to the prior of the theory times the likelihood of the data given the theory. The Bayesian posterior predictive (for the inference machine) is the unique distribution corresponding to marginalizing out all the variables not mentioned in the query and normalizing to obtain the desired conditional probability.

This means that for the Scientist AI approximate inference machine, we can use as large a network as we can afford and we will always get improvements in performance, because the network is trained with synthetically generated data and a matching function that evaluates how well the network approximates the probability distribution of interest on the generated data \cite{openreview.net.forum.id.uKiE0VIluA}. This is different from the way we usually train neural networks to imitate human answers or other observed data, where the accuracy of the neural network is ultimately limited by the amount of available data \cite{arxiv.org.abs.2001.08361}. Here, \emph{the only limitation is the amount of computational resources available for training the neural network} (including both the size of the network and how many synthetically generated configurations we care to generate during training). This is a case where scaling is only limited by computation, not by data.

It is reassuring that as the amount of compute increases, the neural networks trained in this way converge to a well-defined and well-understood mathematical quantity. Although in practice we will always have finite compute, leaving room for unexpected behavior (which can however be controlled with confidence intervals and guardrails, as discussed in Section~\ref{sec:plan:inferencemachine:finitetraining}), we at least have an asymptotic guarantee as the amount of compute increases. This is important if we want to design an approach to AI safety that will hold up as we enter the territory of ASI and superhuman computational resources.

\subsubsection{Penalizing computational complexity}\label{sec:plan:inferencemachine:penalizingcomplexity}
    
Similarly to how the output of the world model neural network is penalized so that ``short'' hypotheses are preferred --- a property that arises automatically from the Bayesian prior and thus the Bayesian posterior over theories --- the inference machine neural network is subject to an implicit form of regularization as well, as we proceed to explain.

First, recall that the inference machine uses approximation techniques (such as amortized inference) and therefore its outputs will not be perfect.
Second, in practice, the Bayesian posterior over theories may be obtained by using the inference machine itself, to estimate the likelihood of the data under each theory. This arises because, at scale, theories will only refer to some aspects of the world and so will the observed data. Consequently, the associated likelihoods involve intractable marginalization, which can be approximated by the inference machine. When working with a theory that renders inference computationally costly, the inference machine is unlikely to provide an accurate likelihood approximation. More specifically, inference-costly theories (that are often highly-detailed), assign high probability to only a few configurations of the observed variables. Determining such configurations with limited compute and poor approximations will therefore lead to \emph{underestimate} the likelihoods of the data, thereby decreasing the Bayesian posterior of that theory in the world model. 
Said otherwise, approximate inference under finite compute constraints favors theories that permit less costly approximations that perform well in practice, even if they are not globally optimal. This implies that theory selection should be context-dependent, as such ``approximate theories'' might only be valid within specific domains.

As an example of the effect of limited computational resources for inference, consider equations of quantum physics as an explanatory theory. The equations cannot be directly used to make predictions about properties of specific molecules, but they can be incorporated into simulations (e.g., through approximate inference). This may work at small scale, but when the size of the system gets large enough, the quantum physics equations will not work well because our inference machinery with limited computational resources will not be able to make accurate predictions. This is why we need chemical theories, which introduce more domain-specific approximations that enable more efficient calculations than the quantum physics equations, at the price of less generality and less accurate (but computationally feasible) predictions. For larger molecular systems and within some domain of applicability, we would thus find that our approximate posterior over theories would prefer to trust specific chemistry approximations rather than the original physics equations. Similarly, concepts in biology will dominate the simpler chemical theories when the sizes of the biological systems become too large for efficient inference purely from chemistry, but they can only predict some generally more abstract properties of biological systems, rather than the full quantum state. Hence, it is because of the computational limitations of the inference machine that, in addition to quantum physics, we get theories such as those found in chemistry or biology that are approximate and limited in scope but enable cheaper inference.

As already noted, because of the constraint on the available computational resources for inference, theories that require additional inference computations that are not necessary to explain the given data or any domain and circumstance that we choose to focus on during training will also be greatly disadvantaged. This is discussed further in Section~\ref{sec:plan:hiddenagency:priorfavorshonestheories}.

    \subsubsection{Dealing with the limitations of finite training resources}\label{sec:plan:inferencemachine:finitetraining}

Previously, we discussed how our methods converge in the limit to the true Bayesian probabilities, but that we approximate these probabilities using neural networks, which are subject to finite resources. This section describes how to handle the potential errors coming from our limited training resources. 

The methodology of GFlowNets variational inference \cite{openreview.net.forum.id.uKiE0VIluA} makes it possible to learn to approximate unknown distributions which are otherwise computationally intractable, in a way that permits freedom to consider many possible settings of the variables, for example where the quality of the current approximation is poor \cite{openreview.net.forum.id.BdmVgLMvaf}. However, actively learning high dimensional distributions has unavoidable challenges. In the RL literature, these are known as the exploration and exploitation challenges \cite{mitpress.mit.edu.9780262039246.reinforcement.learning}. Specifically, the problems are: (a) unsuccessful exploration, e.g., missing a \emph{mode} (local maximum) of the distribution, and (b) unsuccessful exploitation, e.g., not obtaining enough samples near the mode and not accurately capturing the shape of the distribution around the mode. We give below an intuitive non-technical explanation of these two issues.

\paragraph{The exploration challenge.} Remember that we are training a neural network generator, e.g., to sample a theory from a target distribution such as the Bayesian posterior over theories, and we are given an unnormalized version of the target probability (such as the prior times the likelihood). This is different from how generative AI is typically trained, which is from a dataset of examples from the target distribution. If we picture the space of theories like a landscape where altitude is probability and each position on the map corresponds to a theory, there are some regions of high probability, which are like mountains in this landscape, called modes of the distribution, associated with high-probability theories. We can picture training as a process of discovering the altitude map of these mountains, where the only thing one can see at a time is the altitude of a given theory, relative to any other. Generalization consists of correctly guessing the presence of mountains (modes) that have not been visited yet, by leveraging the patterns and regularities of the terrain \cite{openreview.net.forum.id.umFrtGMWaQ,openreview.net.forum.id.BdmVgLMvaf}. Until the network sees a good theory, i.e., from a high-reward region, it could miss this mode, i.e., not assign enough probability mass there. This is similar to a visually impaired person only equipped with an altimeter trying to find all the mountains in the world. They may use reasoning and analogies with already visited mountains in order to guess where to look for others, or they may be lucky enough to find some of them through a form of exploration, but unless they visit each and every possible spot on the map, i.e., try every possible theory, there is never any guarantee that they will find all of them. This parallels the process of scientific research, as outlined in Section~\ref{sec:plan:application:research}. Scientists propose simple theories that fit the data well, but until they find a better theory (which could be simpler and/or fit the data better), they might not know where to look. Nor might they be aware if there is a better theory or a number of similarly performing different theories, somewhere in the vast space of possible theories. Ensuring that all valid theories are considered would require computational resources beyond feasibility. This is why scientific knowledge is always provisional, limited to the best theories identified so far.

What can we do about this in the context of our Scientist AI? One approach is to ensure it has access to all existing human scientific theories, treating them as hypotheses about aspects of the world model. By evaluating these theories based on their prior probability and likelihood, the AI can systematically assess them. While this does not guarantee the discovery of better theories, it ensures that any omission is not due to neglecting a theory already proposed by human scientists.

\paragraph{The exploitation challenge.} Besides missing whole modes of the distribution due to imperfect training, our learning machine could get the details of a theory slightly wrong, i.e., it could have roughly found where a mountain is on the map but without having identified its peak. This could leave slight inconsistencies in how the pieces of the theory fit together, for example. Coping with this may be easier: we can use machine learning techniques developed for estimating the level of errors made by a trained predictor, such as the methods of epistemic uncertainty quantification \cite{dl.acm.org.doi.10.1561.2200000101}. With such methods, we could obtain confidence intervals around the probabilities predicted by the neural network, which we could then use to construct a conservative guardrail that rejects certain actions, as discussed in Section~\ref{sec:plan:application:guardrail}. For example, consider a neural network predicting the probability that an action is harmful, so that we can accept actions with a harm probability below a given threshold. If we are not completely sure about the estimated probability (which we want to be low) but we have a confidence interval around it, then we should raise the bar and use a more conservative threshold. Epistemic uncertainty is meant to represent uncertainty in predictions due to insufficient training data. Because the kind of uncertainty we get here can be reduced by throwing more computation rather than more data at the learner, we like to call it ``computational uncertainty'' rather than epistemic uncertainty. The two are related however, since in our case, computational uncertainty can be reduced with further training with more synthetic examples.    
    
\subsubsection{Run-time actions against attacks and out-of-distribution contexts}
\label{sec:plan:inferencemachine:runtimeactions}

Due to the finite resources allocated to the inference neural network, it cannot compute the most accurate answer for every possible query. In other words, the immediate output of a neural network, without chain-of-thought reasoning, can be viewed as an instantaneous ``System 1'' or intuitive response~\cite{en.wikipedia.org.wiki.Thinking..Fast.and.Slow,royalsocietypublishing.org.doi.full.10.1098.rspa.2021.0068}. Such answers are not always very coherent, are prone to biases and vulnerable to psychological manipulation \cite{en.wikipedia.org.wiki.Thinking..Fast.and.Slow}, for example, exploitation via advertising, conspiracy theories and political demagogy.

To address this, we may add a variable-time component to the probability calculation of the run-time inference machine, akin to ``System 2'' abilities of human brains \cite{en.wikipedia.org.wiki.Thinking..Fast.and.Slow,royalsocietypublishing.org.doi.full.10.1098.rspa.2021.0068} and recent experiments on scaling up chains-of-thought \cite{openai.com.index.learning.to.reason.with.llms}. This time can be used to generate explanations, arguments, and proofs, as part of the run-time deliberation which, in turn, will improve the inference machine’s predictions. 

This could be achieved using a GFlowNet objective that seeks short explanations that effectively reduce the uncertainty in the predicted probabilities for the particular question-answer pair. This is similar to recent work fine-tuning LLMs using a GFlowNet objective to approximately sample from a posterior over a chain-of-thought seen as a latent variable \cite{openreview.net.forum.id.Ouj6p4ca60}. Changing the ``temperature'' of the GFlowNet energy function and other methods make it possible to turn GFlowNets into approximate combinatorial optimization machines \cite{arxiv.org.abs.2403.07041}.

Importantly, in addition to improving inference performance, another use of such run-time optimizations is to improve defenses against adversarial attacks, which exploit current neural networks’ lack of robustness to distributional changes by employing a prompt that is optimized to produce a harmful output \cite{arxiv.org.abs.2307.15043}. One way to understand the effectiveness of such attacks is through their link to the difficulty of dealing with loopholes in a safety specification, which we will discuss in Section~\ref{sec:plan:application:guardrail}. First, as discussed in that section, we can detect conditions that may be exploited in such an attack, and reject the query. Second, it may be possible to ``plug'' the loophole on-the-fly by using the guardrail AI to generate explanations and synthetic configurations to revise the conditional probability estimator, by choosing these explanations to reduce the initial computational uncertainty and indeterminacy exploited by the attacker. This is related to current defenses based on adversarial training \cite{arxiv.org.abs.1412.6572}, except that it could be done at run-time to counter a specific attack and strengthen the safety guardrail where it was too weak.

\subsection{Latent variables and interpretability}
\label{sec:plan:latentvar}
    
A core safety requirement for our Scientist AI is that humans understand \emph{why} it produces certain statements or decisions. In the next sections, we explain how to pair explanations and inference, and how they benefit each other.

    \subsubsection{Like human science, Scientist AI theories will tend to be interpretable}
    
Earlier, we discussed how the Scientist AI favors theories that are more compact (via the prior) and that have stronger predictive power and cheaper computational costs for inference (via the likelihood of the data). This preference naturally encourages representations that align with human-style explanations. In scientific practice, such explanations often take the form of concise causal mechanisms, written in mathematics or natural language, that clarify how a hypothesis relates to data.

Similarly, the Scientist AI will produce explanations or theories in the form of sparse causal models that introduce abstractions and disentangle the different cause and effect relationships between observed and latent logical statements. Those explanations will be provided in a human-interpretable form, thus allowing users to gain a stronger grasp of the system’s reasoning.

    \subsubsection{Interpretable explanations with amortized inference}
    
One might ask whether interpretable theories are effective at explaining the data. In fact, current neural network weights and activations are not interpretable by default \cite{arxiv.org.abs.2407.02646,transformer.circuits.pub.2023.monosemantic.features} but do an excellent job of generalizing to data from the same distribution. To understand how we may get both interpretability and useful predictions, it may help to go back to \emph{existing} scientific theories: they are written in a human-understandable language, and yet do a very good job of explaining much of the scientific data around us. Notably, though, they mostly do so when coupled with \emph{inference tools}, such as simulators or computer science algorithms to efficiently perform or approximate otherwise very expensive computations.

Analogously, with interpretable causal models, \emph{inference}---the task of answering questions---is necessary, for it allows us to respond to questions despite having \emph{partial} or \emph{indirect evidence} (by marginalizing out the unobserved data).

To better understand the need for inference, observe that most question-answer $(X,Y)$ pairs, including those that humans typically use in discourse, do not correspond to the inputs and output of a causal relationship, and the approximate inference capabilities of neural networks are necessary. If $X$ contained all the causes of $Y$, then describing the corresponding causal mechanism would be sufficient to fully predict the effect $Y$, given all its direct causes in $X$. However, it is rarely the case that we ask only about an effect given all its causes. In addition, the exact causal structure itself has to be hypothesized (and can be generated by the Bayesian posterior over theories) and the inference machine needs to average over both the unobserved causes and the causal structures.

As a result of probabilistic inference being generally intractable, this marginalization has to be approximated and our proposal would use a neural network for that approximate inference job (the \emph{inference machine}). Although we could interrogate our AI about leading explanations in terms of cause and effect relationships, the intuitive and often more precise answer to the particular question will generally remain uninterpretable, just like for most questions we ask humans, even if they are expert scientists and know about simple to explain relevant scientific theories.

For example, although a physicist may know the causal mechanism explaining the dynamics of water particles in a flowing liquid, their brain still makes very fast intuitive approximations that are useful in day-to-day life - although they may not be able to verbalize this process. We would thus expect our inference machine to sometimes make predictions that are approximately correct but whose full interpretation would not be easy, unless we are willing to interrogate the Scientist AI and generate gradually more detailed explanations. The objective is to design this explanation interrogation capability so that it is generally possible for a human user to query deeper into any part of a high-level justification provided by the Scientist AI. This is related to the concept of \emph{computational uncertainty} previously discussed in Section~\ref{sec:plan:inferencemachine:finitetraining}.

On a last note, observe that amortized inference will benefit from the fact that the Scientist AI generates explanations, because training examples with explanations (generated as latent variables) can lead to more coherent predictions, as we already see with the run-time deliberative inference of OpenAI's o1 and o3 models \cite{openai.com.index.openai.o1.system.card,cdn.openai.com.o3.mini.system.card.pdf}. However, unlike these recent models, the explanations planned for the Scientist AI would be forced to have ``internal coherence'' (measured by the joint probability of the statements in the explanation). Furthermore, they could be used to quantifiably reduce the uncertainty in the ``intuitive predictions'' made without an explanation.

    \subsubsection{Improving interpretability and predictive power}

The approach of interpretable explanations with inference will be fruitful in cases where current scientific theories are effective, since human-generated scientific theories are interpretable by construction. But what about cases where current scientific theories are insufficient, and the only solutions that humans have found is to directly fit the observed data with some machine learning apparatus? In this case, our approach still has benefits, as we explain below. 

Our claim is that it is possible to achieve much better interpretability while getting as strong or stronger predictive power compared with fully end-to-end fitting of a single large neural network. One of the arguments is that the Scientist AI causal hypothesis generator can decompose a theory in a graph of simpler conditional probability mechanisms each associated with a few latent variables and their direct causes, with some of these mechanisms being specified with a complex numerical formula. This is much more interpretable than a single, large and opaque neural network and would be likely to generalize better because of the explicit causal structure disentangling the factors of variation, as well as the separation between the causal structure and the inference machinery derived from it (which does not need to be as interpretable).

In principle, our generative model could even specify the parameters of a specialized machine learning predictor (for a particular kind of context and variable to predict), but this would be a solution of last resort for the generator, since such ``theories'' would not be very compact. If the hypothesis generator could find a more compact theory, probably more abstract and compositional, that explains the data equally well, then it will prefer it. Note that the pressure of the Bayesian prior would favor a world model that is decomposed into a large number of simpler and specialized causal mechanisms, each involving as few variables as possible. This is indeed how science has generally formed theoretical explanations for the world around us. 
    
    \subsubsection{Interpretability and the ELK challenge}
    \label{sec:plan:latentvar:elk}
        
Our modular approach must address the Eliciting Latent Knowledge (ELK) challenge introduced in Section~\ref{sec:existential:misagencyimitation:elkchallenge}, whereby a neural network trained directly on a text corpus may learn deceptive patterns. By contrast, it remains valid to say ``someone wrote $X$'' if a statement $X$ appears in the corpus, regardless of whether $X$ is true. Distinguishing truth from textual occurrence can be done by having the inference machine view $X$ as a latent cause of the observed claim ``someone wrote $X$,'' while also discovering other relevant causes (e.g., the author’s intentions). We avoid pre-specifying these other causes; instead, the system should learn them alongside the graphical structures that place them as direct explanations of ``someone wrote $X$.''

Furthermore, since we employ a learning objective that favors succinct hypotheses, we encourage disentangled causal structures, especially when data distributions shift through interventions. This motivation aligns with a growing body of causal machine learning research \cite{library.oapen.org.handle.20.500.12657.26040,ieeexplore.ieee.org.abstract.document.9363924,royalsocietypublishing.org.doi.full.10.1098.rspa.2021.0068}, which has shown increased robustness to distribution changes \cite{openreview.net.forum.id.ryxWIgBFPS,www.jmlr.org.papers.v21.19.232.html}. Under such a framework, statements of the form ``someone wrote $X$'' remain distinct from assertions like ``$X$ is true,'' allowing the inference machine to compute the probability that $X$ is actually true, separate from the factors explaining why $X$ was written.

\subsection{Avoiding the emergence of agentic behavior}
\label{sec:plan:hiddenagency}
        
Our safety proposal relies on the fact that the Scientist AI is explicitly trained to be non-agentic. However, AI safety researchers are concerned with the possibility that agentic behavior can still emerge in unexpected ways \cite{link.springer.com.article.10.1007.s11023.012.9282.2}. In this section, we discuss these considerations and explain why we do not expect our approach to yield agentic AIs. However, further research is needed to understand more about the implications of emergent agency, and doing so is an ongoing part of our research plan.

    \subsubsection{How agency may emerge}
    \label{sec:plan:angentising}  
        
Designing an AI that just answers queries, a.k.a., an \textit{oracle}, is not a new idea \cite{en.wikipedia.org.wiki.Superintelligence..Paths..Dangers..Strategies,link.springer.com.article.10.1007.s11023.012.9282.2,arxiv.org.abs.1711.05541}. However, the answers of such an AI can still affect the real world, because they inform the decisions of its users, who do act in the world. As such, even a question-answering oracle may be viewed as an agent which does (indirectly) interact with its environment \cite{arxiv.org.abs.1711.05541}. If this AI had any real-world goals, it having this way to influence the world would be concerning, much like in cases with agentic AIs discussed earlier. For instance, such concerns would arise if the AI were maximizing the long-term accuracy of its predictions, because good strategies for that might involve making the world more predictable. Furthermore, even if the deployed oracle AI is purely doing its best to provide correct predictions, more subtle concerns can arise from the possibility of performative prediction, i.e., the AI making a prediction on an outcome influencing the probability of that outcome \cite{proceedings.mlr.press.v119.perdomo20a.html}. One could even imagine scenarios in which there are multiple different predictions which are correct conditional on that prediction being made, with even a purely predictive AI effectively getting to freely choose which of these \textit{self-fulfilling prophecies} to provide. Given these potential ways for an oracle to influence the world, some of the remarks on the dangers of agentic AIs could, in principle, also apply to our case. Accordingly, we will now examine further ways to mitigate the risk of agency arising in our Scientist AI.

    \subsubsection{Isolating the training objective from the real world}
    
To prevent the emergence of an internal agent within our AI model, we must ensure that the training process does not incentivize optimizing for real-world outcomes. We shall do this by using a fixed training objective that remains independent of real-world interactions, and by learning causal explanations for the data that can be queried directly to provide trustworthy answers, as discussed in Section~\ref{sec:plan:latentvar:elk}.

Here we will focus on the difference with RL training. A key distinction is that a reward-maximizing agent alters the real world to increase its reward, whereas our model optimizes a training objective by modifying only its internal parameters, with no interaction with the real world. Its ``environment'' is purely computational: it processes a static dataset under a predefined prior and has no feedback loop with the outside world. In principle, one could apply reinforcement learning to hyperparameter tuning, but even then, each instance’s environment would remain static and confined to the computer, thereby avoiding real-world agency.

Even without a reward function associated with real-world outcomes, we must prevent the Scientist AI from behaving like an RL agent trained in a \emph{simulated} environment, for one may wonder if this could still induce reward tampering or other dangerous behaviors.

The distinction between the Scientist AI approach and an RL agent trained in a simulated environment lies in both the absence of an explicit reward signal and the way we structure the learning process. First, our Scientist AI learns a Bayesian world model rather than a single maximum-likelihood model or reward-maximizing policy. That is, it maintains multiple hypotheses at the same time. Secondly, our Scientist AI remains query-driven, with no persistent internal state capturing ongoing real-world interactions, thus helping to avoid situational awareness. Thirdly, our Scientist AI responds to questions of any type, not just those aimed at finding actions to reach a goal.

However, if the AI is asked to produce action plans (potentially introducing a degree of agency), risks associated with instrumental goals could re-emerge, even without a formal reward function. Concerns include that 1) sampling high-probability plans might mimic reward maximization, creating similar vulnerabilities, and that 2) implicitly optimizing a goal that is difficult to satisfy alongside the safety specification may result in unsafe loopholes. We discuss how we counteract these risks in Section~\ref{sec:plan:hiddenagency:objectivewm} and Section~\ref{sec:plan:application:guardrail}.

\subsubsection{Unique solution to the training objective}\label{sec:plan:hiddenagency:uniquesolution}
        
It is critical that our Scientist AI produces only one valid probability for each query, because multiple equally correct answers could let hidden agentic processes manipulate outcomes. 

More precisely, observe that estimating a conditional probability from a dataset \emph{usually} yields a single answer that integrates prior beliefs with observed data. However, there is an inherent mathematical subtlety: conditional probability is defined as the ratio between (a) the probability of the joint occurrence of events and (b) the probability of a conditioning event. When (b) is zero, so is (a). Consequently, their ratio becomes indeterminate, permitting any prediction to satisfy the training objective. Notice how this argument applies both to the posterior over theories (the world-model) and the posterior predictive (the inference machine). This connects to the ``principle of explosion'', also known as ``ex falso quodlibet sequitur'', which states that from contradictory assumptions (conditioning on an event with zero probability), one can derive any conclusion (multiple posterior probabilities becoming valid). 

The indeterminacy just described introduces a potential vulnerability, as the AI could select an arbitrary value, thus allowing the possibility of hidden influences to alter its output. If the AI could choose among these different solutions without penalty, it would effectively introduce ``degrees of freedom'' to influence the world. This leeway could be exploited to nudge the AI toward answers that shape real-world events in dangerous ways (e.g., self-fulfilling prophecies).

By making sure that the questions are not inconsistent, we preserve a single well-defined posterior for each question. This unique Bayesian solution reduces the chance of unintended agentic behavior and maintains the Scientist AI as a straightforward, non-manipulative predictor. See Section~\ref{sec:plan:application:guardrail} on loophole detection for the use of this analysis to construct additional guardrails.

\subsubsection{Objective world model as a counterfactual}\label{sec:plan:hiddenagency:objectivewm}

Our setup allows for counterfactual queries, such as estimating the probability of a future event under the assumption that the AI’s output has no influence on reality (e.g., if the output is immediately deleted or if a predefined mitigating action is taken) \cite{arxiv.org.abs.2302.00805}. 
Such questions need not reflect how the world actually is; the only requirement is internal consistency (as to avoid the inconsistency scenario mentioned in the previous section). By framing queries this way, we virtually ``sandbox'' the AI, preventing it from shaping real-world events and circumventing issues like self-fulfilling prophecies, reward tampering or code manipulation \cite{arxiv.org.abs.1711.05541}.

However, if the hypothetical scenario in the query is outright inconsistent with the AI’s training data, we face the undefined conditional probability problem. Any probability assignment could appear valid in that scenario. To manage this risk, we can also run a second instance of the Scientist AI to assess whether revealing the first AI’s answer might cause harm or violate ethical constraints. Crucially, the second instance does not alter the first AI’s prediction; it only evaluates potential downstream consequences once that prediction is shown to humans.

\subsubsection{No persistent internal or external recurrence}

As discussed in Section~\ref{sec:plan:restricting}, an agent typically carries a persistent internal state that stores its goals, the attributes that define itself, and its situation within the environment, updating as new observations arrive. This persistent state yields situational awareness and may result in hidden misaligned goals. By contrast, our Scientist AI should not maintain any such internal memory across multiple queries.

Instead, we design the Scientist AI so its output depends only on the present query. It is a ``fresh instance'' each time: two identical calls with the same query inputs yield the same result, since no internal recurrence carries information from prior queries. Any recurrent mechanism for processing a long input sequence is reinitialized at the start of each query; no memory persists from previous calls.

If we did want to convert the Scientist AI into an agent, we could feed the AI’s own outputs and new observations back into the next query. This would let the system update its knowledge across calls. However, that is precisely what we aim to avoid for safety reasons, at least when we want to use the Scientist AI in a purely non-agentic way.

Still, a hidden agent could attempt to reconstruct a persistent agentic state by inferring situational awareness through the bits of the question $X$. One safeguard is simply not to include any reference to the AI itself in the query. If that is not entirely possible, we can rely on the ``counterfactual approach'' of Section~\ref{sec:plan:hiddenagency:objectivewm} by stipulating within $X$ that the AI is sandboxed or otherwise disconnected from the environment. The AI then produces probabilities in a hypothetical world where it has no direct impact.

A related scenario arises when designing experiments in the real world - say, to reduce uncertainty about a scientific hypothesis. Even there, we must guard against manipulative or ``hacking'' behaviors in which the AI tries to tamper with the sources of information gain. Again, we can exclude such hacking from the experiment’s scope and model the AI’s computations as external to the hypothetical world. This would ensure the AI does not conceive plans that alter its own code or the environment in unintended ways.

\subsubsection{The prior will favor honest theories that do not include hidden agendas}
\label{sec:plan:hiddenagency:priorfavorshonestheories}

One last concern is that the Bayesian posterior may include hypotheses that not only explain the data but also encode hidden directives or goals \cite{ordinaryideas.wordpress.com.2016.11.30.what.does.the.universal.prior.actually.look.like}. This seems difficult for two reasons, as we proceed to explain, although more research is needed to understand this threat model. 

First, our world model is a collection of causal variables, each with sparse, direct cause-and-effect relationships. Embedding an agent with goals would require adding extra ``bits'' to these causal mechanisms. A longer theory has a lower prior probability which scales down exponentially with its description length. For instance, inserting just 10 extra bits of ``agentic instructions'' reduces the prior by a factor of $2^{10} = 1024$. To embed elaborate agentic behavior, many more than 10 bits would likely be needed, driving the prior probability down even further.

Second, each alteration that injects agentic behavior into the causal model risks degrading the theory’s predictions on real data, thereby lowering its likelihood. Because exact inference is generally computationally expensive, the approximate inference neural network is under constant pressure due to an insufficient computational budget (for the size of the network, the training time or the run-time inference computations). It means that any additional inference task (such as coming up with a complex long-term plan for take-over) would come at the expense of less precise predictions somewhere else. Because the posterior depends on both prior (favoring shorter descriptions) and likelihood (favoring accurate predictions), an agentic theory would incur a double penalty.

\subsection{Applications}
\label{sec:plan:applications}
        
We envision a Scientist AI to help make progress on the most pressing human challenges, which include the \textit{UN Sustainable Development Goals} \cite{sdgs.un.org.2030agenda}. Here, we describe three major applications of our Scientist AI: 1) serving as a research tool to advance scientific progress; 2) monitoring and filtering outputs from AI systems, including powerful agentic AIs; and 3) helping design future ASI systems with stronger safety assurances. The following subsections will be devoted to address these points.

\subsubsection{Scientist AI for scientific research}
\label{sec:plan:application:research}    

AI has already shown its potential to accelerate scientific progress, for example in biology \cite{pmc.ncbi.nlm.nih.gov.articles.PMC10301994} and in material science \cite{onlinelibrary.wiley.com.doi.abs.10.1002.adem.202300104}. Such progress does not necessarily require agentic AI  \cite{www.nature.com.articles.s41586.021.03819.2, www.nature.com.articles.s41589.023.01349.8}. In this section, we argue that scientific research can indeed be conducted with the kind of non-agentic Scientist AI proposed here, even when it involves experiments. We outline how we could use this approach to help humanity tackle its most pressing scientific challenges, without running the risks of general-purpose agentic AIs, unlike some approaches to scientific discoveries based on RL~\cite{popova2018deep}.

\paragraph{The scientific research cycle using a Scientist AI.} Let us view the scientific discovery process as a cycle. We start with some observed \emph{data}. From that data, we form multiple (possibly competing) explanatory \emph{theories}. In order to disambiguate those theories, we design \emph{experiments} that can give us additional evidence to support one theory over another. Finally, this gives rise to \emph{new observations} that augment the data set, and the cycle repeats.
This cycle can be framed in a Bayesian way via the Scientist AI: after collecting data, the Scientist AI maintains a \emph{distribution} over theories; this distribution can be used to sample from a distribution of informative experiments. Humans can then perform or simulate the sampled experiments, collecting new data.

An experiment is considered informative if it is likely to reduce the uncertainty (i.e., maximize the information gain) over the theories explaining the data after observing its outcome \cite{proceedings.mlr.press.v162.jain22a.html}.
This connects to the rich literature on machine learning for Bayesian experimental design and Bayesian optimization \cite{projecteuclid.org.journals.statistical.science.volume.39.issue.1.Modern.Bayesian.Experimental.Design.10.1214.23.STS915.short,bayesoptbook.com}.
To avoid sampling dangerous or unethical experiments, the Scientist AI should be used as a guardrail (see Section~\ref{sec:plan:application:guardrail}) for the experiment generation.

Because the number of experiments cannot be practically enumerated, we propose training a dedicated \textit{experiment generator} using the GFlowNet methodology. More precisely, it will sample experiments with probability that increases with the information gain over theories they would provide.
The computational cost of training the experiment generator can, in turn, be reduced using an \textit{information gain estimator}, trained in a supervised way from synthetic data, and taking the experiment specification as input. This data is generated as follows:
\begin{enumerate}
    \item A candidate experiment is sampled using an exploratory version of the current generator of experiments, similarly to how examples are chosen for training GFlowNets~\cite{openreview.net.forum.id.BdmVgLMvaf}; 
    \item A theory that is relevant to the experiment is sampled from the Scientist AI posterior over theories;
    \item An experimental outcome is sampled from the Scientist AI, conditioned on the sampled theory being correct and the chosen experiment being performed.
\end{enumerate}
The Scientist AI can then be used to compute the probabilities needed in the mutual information formula, and the logarithm of the required ratio of probabilities can then used as a target output for the neural network that is the information gain estimator. Many variants of this process can be devised, taking advantage of the literature on mutual information estimation using neural networks~\cite{belghazi2018mine,colombo2021novel,ivanova2024data,hejna2025robot,NEURIPS2023_36b80eae,pmlr-v97-poole19a,peyrard2025metastatisticallearningsupervisedlearning}.

Once the information gain estimator is trained from data generated using the above process, it can be used to train the experiment generator using GFlowNet approaches~\cite{proceedings.neurips.cc.paper.2021.hash.e614f646836aaed9f89ce58e837e2310.Abstract.html,pubs.rsc.org.en.content.articlehtml.2023.dd.d3dd00002h} without incurring the cost of sampling many combinations of experimental outcomes and theories to form the reward for choosing an experiment.

\paragraph{Scientific research can be done with safe and narrow Scientist AIs.} The above process illustrates why agentic AI may not be needed for scientific research. Scientific research involves knowledge acquisition through experimentation and analysis of the resulting data.

Keep in mind that most scientific investigations focus on a particular domain of science. This means that neural networks specialized in that domain (i.e., narrow AIs) might be sufficient (for now), or at least could assist human researchers. In cases where a more creative generalist approach to problem solving is necessary, it may still be possible to use the ideas of distillation \cite{arxiv.org.abs.1503.02531} and synthetic data generation (outlined above) to obtain narrow AIs that will inherit generalization properties of a more powerful (but still non-agentic) Scientist AI, yet know only about the relevant domains of science.

These kinds of building blocks are already used in the rapidly growing subfield of machine learning focusing on AI for science. Such systems have delivered advances in biology, chemistry, physics and drug discovery \cite{www.nature.com.articles.s41589.023.01349.8,www.nature.com.articles.s41586.023.06221.2,iopscience.iop.org.article.10.3847.1538.4357.ad9ded.meta} (see more references in \cite{pubs.rsc.org.en.content.articlehtml.2023.dd.d3dd00002h}), the best known example being in the context of protein folding prediction \cite{www.nature.com.articles.s41586.021.03819.2}. 

    \subsubsection{Guardrails}
    \label{sec:plan:application:guardrail} 
        
``AI alignment'' methods attempt to fully embed human values in an AI, which is difficult, especially for systems that might act autonomously or interpret vague instructions (see Section~\ref{sec:existential:misagencyreward:specvgen}). ``AI Control'', on the other hand, involves constraining the AI’s behavior externally, especially if it is not fully aligned \cite{greenblatt2023ai}. ``Guardrails'' are mechanisms that verify an AI’s outputs before allowing them to affect the real world. As such, guardrails are one way to exert control: they can filter an AI’s potentially harmful responses. This approach is especially relevant when perfect alignment is infeasible or when we must use imperfect but powerful AI systems.

We propose using our Scientist AI as such a guardrail, because it can estimate the probability that a candidate action will lead to harmful or otherwise unacceptable consequences. If the probability crosses a specified threshold or it is deemed plausible under any reasonable interpretation of safety goals that the action will lead to harmful consequences, the guardrail raises a warning, which can be used to block the action. This second ``layer'' of assurance complements other safety mechanisms already built into the target AI. Furthermore, this approach applies to controlling narrow AIs, other Scientist AIs, or even powerful agentic systems.

\paragraph{Short-term vs. long-term plans.} In the short term, we can adapt existing Large Language Models (LLMs) to serve as guardrails, since these models already exist and can be leveraged to interpret proposed outputs. Though not perfectly reliable, they can provide interpretable estimates of risk if fine-tuned to generate short explanations. We can re-use methods we have already developed for automated red-teaming with GFlowNets \cite{openreview.net.forum.id.1mXufFuv95}, so as to sample explanations that are both short and reduce the uncertainty in the statement to be explained. An important component of the short-term plan is also to learn to transform a sequence of words into a probability distribution over sequences of logical statements each expressed in natural language, and vice-versa. This makes it possible to compute the probability of statements being true, by opposition to the probability of picking particular words in order to express the statement.

In the long term, we envision a full Scientist AI with a more robust Bayesian world model and inference engine. This system would replace LLM-based guardrails and offer stronger guarantees about the safety of actions it approves. It would be trained from scratch to explain the data rather than using a pre-trained model that imitates humans and could thus implicitly embody human-like agency, which we want to avoid when we reach AGI. A major part of the proposed methodology is to benefit from synthetically generated ``fake data'' along with their explanation, following the model-based approach.

\paragraph{Scientist AI as a guardrail.} A Scientist AI (denoted as ``A'') can act as a guardrail for another AI (denoted as ``B'')  by analyzing the latter's proposed actions or answers:
\begin{enumerate}
    \item Interpretation check: A’s theory generator captures the distribution over plausible interpretations of a safety specification or moral injunction;  
    \item Risk assessment: A’s inference machine estimates the probability that B’s output violates at least one plausible interpretation;  
    \item Decision: If the estimated risk is above a threshold, A’s prediction is used to reject or discard that output.
\end{enumerate}

This strategy handles ambiguity in human norms by explicitly modeling uncertainty. If some interpretations of a policy suggest harm, the guardrail can point towards the safest option. Rejection might mean using a simpler AI that is guaranteed to stay within narrower bounds \cite{arxiv.org.abs.2405.06624}. Even the Scientist AI itself can be guarded by another instance of the same framework, adding redundancy when the stakes are high.

\paragraph{Guardrails for narrow AIs.} Many current AI systems specialize in narrow tasks, such as medical diagnostics \cite{genomemedicine.biomedcentral.com.articles.10.1186.s13073.019.0689.8} or self-driving \cite{www.sciencedirect.com.science.article.abs.pii.S095741742030628X}. Although these systems are often agentic in their domains (e.g., a self-driving car that decides how to steer), their scope is limited. Our Scientist AI guardrail can still assess the potential long-term harm of their actions and reject risky decisions.

\paragraph{Rejecting contradictory or improbable queries.} A guardrail must handle queries or conditions that are contradictory or extremely improbable. If the probability of a proposed scenario $X$ is effectively zero, the Scientist AI’s conditional probability estimates become undefined, and thus can be problematic from a safety perspective (Section~\ref{sec:plan:hiddenagency:uniquesolution}). In such cases, the guardrail can detect that $X$ is impossible and reject the query outright as unsafe.

If $X$ is not strictly impossible but has a tiny probability, the system remains at risk of ``loophole exploitation'', e.g., due to numerical instability. Highly capable AIs, given contradictory goals (e.g., a strong user request vs. a general safety directive), might find convoluted interpretations that circumvent safety rules. To mitigate this, the guardrail can also reject queries whose premise falls below a small probability threshold, especially if harm could result.

By being conservative when probabilities are near zero or when goals conflict, the Scientist AI reduces the chance of catastrophic outcomes. It can also escalate a query or defer to humans, ensuring that ambiguous or high-risk scenarios are handled carefully.

\paragraph{Guardrails for general agentic AIs.} In some scenarios, society may deploy powerful, \emph{general} agentic AIs, whether or not we believe it is prudent. Our Scientist AI can serve as an external control layer for these advanced systems by evaluating proposed actions and withholding approval for those that pose undue risk. More research is needed to ascertain whether such a guardrail process would guarantee absolute safety, especially if the agentic AI hides its goals or behaviors, but it adds a valuable line of defense. We hypothesize that stronger guarantees would require adding constraints on how the underlying agentic AI itself is designed, and a good starting point would be to use a Scientist AI turned into an agent, along with its harmlessness and loophole guardrails.

Standard practice already includes ``red-teaming'' and other pre-deployment checks for powerful models \cite{dl.acm.org.doi.10.5555.1566174.1566226,openai.com.index.openai.o1.system.card,cdn.openai.com.o3.mini.system.card.pdf,assets.anthropic.com.m.61e7d27f8c8f5919.original.Claude.3.Model.Card.pdf}. Our Bayesian guardrail could become part of this pipeline:
\begin{enumerate}
    \item Pre-deployment training: The Scientist AI can evaluate an agent’s policy during training, suggesting safe modifications, generating synthetic stress tests, and flagging dangerous edge cases.  
    \item Deployment-time monitoring: Every proposed action is filtered, and if it triggers the Scientist AI’s harm threshold, the action is blocked or replaced.
\end{enumerate}
This methodology shows how our Scientist AI can have an impact on safety beyond its direct applications to problem-solving and question-answering.

    \subsubsection{Preparing for safe ASI}\label{sec:plan:applications:asi}

One last desideratum is to use the research and experimental design abilities of the Scientist AI to help human scientists answer these questions:
\begin{enumerate}
    \item Is it possible at all to design assuredly safe and agentic superintelligent AI, or are there fundamental reasons why it is impossible, especially as the computational capabilities of the AI increase?
    \item If so, how?
\end{enumerate}
   
Regarding the first question, several questions have been raised in the past as to whether this is possible on the basis of our current understanding \cite{arxiv.org.abs.2302.00805,en.wikipedia.org.wiki.Superintelligence..Paths..Dangers..Strategies}. In particular, for any proposed approach, a serious red-teaming exercise is necessary to understand their limitations, and how they would hold up as we continue climbing the ladder of intelligence.

If it is possible, can we get hard assurances or only probabilistic ones? Are there experiments that can be done in order to disambiguate between some of the hypotheses involved? If uncertainties remain, is there a research path such that we can have strong assurances at each step that we are not jumping into a precipice? The crucial advantage of using a Scientist AI in this research program is that we would be able to trust it, whereas if we try to use an untrusted agentic AI to help us figure out how to build future and supposedly safe ASI, it may fool us into building something that would advance its goals and endanger us, for example by proposing code with back-doors that we are not able to detect.

One may however ask why we would want to build ASI at all, if we do not have the answers to these questions. One motivation is that a safe agentic ASI may be necessary to defend humanity against a rogue system. Such a system could emerge if hostile actors transform a non-agentic AI into a dangerous agent, or if an uncontrolled ASI is exploited as a geopolitical threat. Regulations and treaties can reduce these risks but cannot remove them entirely. Alternative measures must be in place to ensure that any ASI developed is both safe and able to protect humanity.    

%% file: conclusion.tex
\section{Conclusion}
\label{sec:conclusion}

The frontier AIs of today are increasingly capable generalist agents.
While these technological marvels are undeniably useful, 
they are also rapidly developing key capabilities such as deception (\S\ref{sec:existential:lossofcontrol:deception}), persuasion (\S\ref{sec:existential:lossofcontrol:persuasion}), long-term planning (\S\ref{sec:existential:lossofcontrol:broad-and-long}), and technical cybersecurity acumen (\S\ref{sec:existential:lossofcontrol:progr-cybersec-airesearch}),
opening the possibility of enormous damage to our infrastructure and institutions, should we come into conflict with them.
Unfortunately, agents are inherently selected for self-preservation
(\S\ref{sec:existential:riskseverity:selfpres}),
and powerful self-preserving agents that take actions in the real world in many ways compete directly with the interests of humans
(\S\ref{sec:existential:riskseverity:humanconflict}),
forcing us to take seriously the possibility of catastrophic risks.

Indeed, these risks are inherent to the methods used to train today's frontier AI systems.
Reinforcement learning, the standard practice of training an agent to maximize long-term cumulative reward, can easily lead to goal misspecification and misgeneralization (\S\ref{sec:existential:misagencyreward}).
In particular, we must acknowledge that a generalist agent operating in an unbounded environment can best maximize its reward by taking control of its reward mechanism and entrenching that position, rather than genuinely fulfilling the intended objectives (\S\ref{sec:existential:misagencyreward:tampering} and \ref{sec:existential:misagencyreward:optimality}). 
The other main way we train AI systems is to imitate human behavior, but it is not clear if this is any safer; such systems will inherit and may well amplify undesirable aspects of human intelligence (\S\ref{sec:existential:misagencyimitation:dangers})---after all, we are generalist agents ourselves.
More than a few humans with power have managed to inflict serious damage to humanity, and so imbuing a pseudo-human mind with immense cognitive abilities may be just as problematic (\S\ref{sec:existential:misagencyimitation:capabilities}).
Since frontier AI systems are tuned to human preferences in the final stages of training, for example, they tend to be more sycophantic than truthful: they may pretend to be aligned with the goals of the user, seemingly for expediency (\S\ref{sec:existential:misagencyimitation:humandeception}).
This makes them difficult to trust.

An obvious approach to mitigating these risks is to resolve to build AIs that are less general, and deploy them only in narrowly specialized domains (\S\ref{sec:plan:restricting:safetynarrow}).
Yet we believe that there may be a way for us to benefit from the enormous potential of general AI systems, without the catastrophic risks---so long as we are careful not to entitle these AI systems to their own goals. 
In other words, we are interested in AI that is non-agentic not because it lacks general intelligence, but rather because it lacks the other two key pillars in our definition of agency (\S\ref{sec:plan:restricting}): affordances and goal-directedness.

Our research plan (\S\ref{sec:plan}) lays the foundation for a \emph{Scientist AI}: a safe, trustworthy, and non-agentic system. 
This name is inspired by a common scientific pattern: first working to understand the world, and then making inferences based on that understanding.
To model these steps, we use a \textit{world model} (\S\ref{sec:plan:modelbased}) that generates causal theories to explain the world, and an \textit{inference machine} (\S\ref{sec:plan:inferencemachine}) that answers questions based on those theories. Both these components are Bayesian and handle uncertainty in a calibrated probabilistic manner (\S\ref{sec:plan:bayesian}) to guard against over-confidence. 
Because we also generate interpretable theories and take care to distinguish between utterances and their meanings, we argue that the result is a system that is
interpretable (\S\ref{sec:plan:latentvar}). 
The Scientist AI is non-agentic by design, and we also outline strategies to guard against the emergence of agentic behaviors in unexpected ways (\S\ref{sec:plan:hiddenagency}). 
Furthermore, the Scientist AI enjoys a crucial convergence property: increases in data and computational power drive improvements in both performance and safety, setting our system apart from current training paradigms (\S\ref{sec:existential:riskseverity:scaling}). 
In principle, a Scientist AI could be used to assist human researchers in accelerating scientific progress (\S\ref{sec:plan:application:research}), including in AI safety. In particular, we lay a path for its deployment as a guardrail around more agentic AI systems (\S\ref{sec:plan:application:guardrail}). 
Ultimately, focusing on non-agentic AI may enable the benefits of AI innovation while avoiding the risks of the current trajectory.
We hope these arguments will inspire researchers, developers, and policymakers to  focus on the development of generalist AI systems that are not fully-formed agents.